%% file: acl_latex.tex
\documentclass[11pt]{article}

\usepackage[preprint]{acl}
\usepackage{svg}

\usepackage{times}
\usepackage{latexsym}

\usepackage[T1]{fontenc}

\usepackage[utf8]{inputenc}

\usepackage{microtype}

\usepackage{inconsolata}
\usepackage{pgfplots}
\pgfplotsset{compat=1.17}
\usepackage{graphicx}
\usepackage{amsmath}
\usepackage{amssymb}
\usepackage{booktabs}
\usepackage[table]{xcolor}
\usepackage[most]{tcolorbox}
\usepackage{multirow}
\usepackage{array}
\usepackage{enumitem}
\usepackage{float}
\usepackage{svg}
\definecolor{promptbg-instgen}{RGB}{255,248,230}
\definecolor{promptframe-instgen}{RGB}{217,119,6}
\definecolor{promptbg-zero}{RGB}{240,249,255}
\definecolor{promptframe-zero}{RGB}{14,116,144}
\definecolor{promptbg-image}{RGB}{239,246,255}
\definecolor{promptframe-image}{RGB}{37,99,235}
\definecolor{promptbg-instr}{RGB}{255,251,235}
\definecolor{promptframe-instr}{RGB}{180,83,9}
\definecolor{promptbg-hybrid}{RGB}{240,253,244}
\definecolor{promptframe-hybrid}{RGB}{5,150,105}
\definecolor{promptbg-flip}{RGB}{245,243,255}
\definecolor{promptframe-flip}{RGB}{109,40,217}
\tcbset{promptbox/.style={
  enhanced, breakable, sharp corners,
  boxrule=0.5pt, left=4pt, right=4pt, top=4pt, bottom=4pt,
  fonttitle=\bfseries, coltitle=white
}}
\title{Instruction Distillation: Text Instructions as Visual Examples}

\author{Hardik Jindal \\
    IIT Kanpur \\
  \texttt{hardikj22@iitk.ac.in} \\\And
  Soumyabrata Pal \\
  Adobe Research\\
  \texttt{soumyabratap@adobe.com} \\\And
  Sayak Ray Chowdhury \\
    IIT Kanpur \\
  \texttt{sayakrc@iitk.ac.in} \\}

\begin{document}
\maketitle

\begin{abstract}
Visual in-context learning (ICL) with multimodal large language
models (MLLMs) is effective for fine-grained visual classification,
but each retrieved image example consumes several hundred context
tokens, making large-$K$ settings prohibitively expensive at
inference scale. We propose \textbf{instruction distillation}: an
offline procedure in which the MLLM itself generates, for each
individual training image, a structured identification instruction
encoding general appearance cues, features that differentiate the
class from visually similar ones, and a common confusion point.
Unlike prior work that produces a single description per class, our
instructions are generated per training image, preserving the
intra-class visual diversity that per-class descriptions collapse.
At inference time, we study five configurations sharing a single
CLIP retrieval index: zero-shot, image ICL, instruction-only ICL,
and two hybrid variants in which retrieved neighbors are split
between images and instructions. Across seven fine-grained
benchmarks and two MLLM backbones, instruction based pipelines match, or
exceeds image ICL at $K{=}1$ and
reduces per-query tokens by $2.9\times$ and inference latency by $3.3\times$ at $K{=}5$. Hybrid configurations further show that visual and
textual ICL signals are complementary, images give visual patterns to learn and see, while instructions give explicit rules and logic. When both of these are provided, the quality of context improves, which is noticeable in the performance.
\end{abstract}

\section{Introduction}

\begin{figure}
    \centering
    \includegraphics[width=1\linewidth]{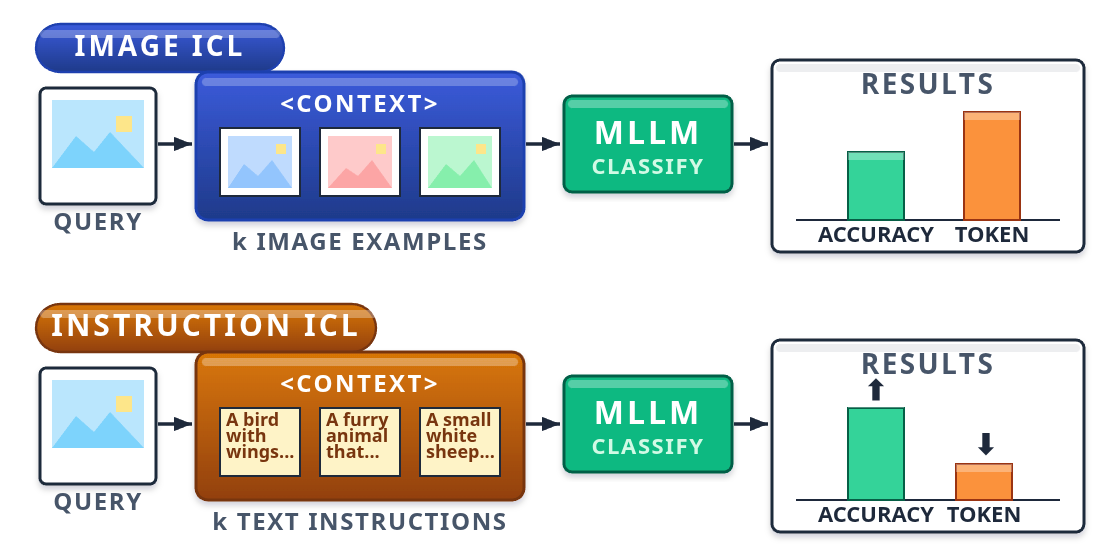}
    \vskip -3mm
    \caption{\textbf{Instruction Distillation.} Images are replaced by text-based instructions and passed as context during inference. It results in an improved accuracy score with reduced token consumption.}
    \label{fig:instr}
\end{figure}
Multimodal large language models (MLLMs) can perform visual classification in a few-shot setting: given a handful of image–label pairs as in-context examples, they classify novel images substantially more accurately than under zero-shot prompting~\citep{alayrac2022flamingo,zong2024vlicl}. This visual in-context learning (ICL) paradigm is particularly effective for fine-grained recognition, where categories exhibit high inter-class visual similarity, and the model must attend to subtle discriminative cues, such as surface texture, structural configuration, or preparation style.

The standard recipe is retrieval-augmented: for each test image, the
$K$ most similar training images are retrieved via CLIP~\citep{radford2021clip}
and placed in the model's context. This works~\citep{jiang2024manyshot,
liu2022makesgoodexamples}, but carries a hidden cost that becomes
prohibitive at scale. Each image consumes roughly 550 tokens in the
form of visual patch embeddings, so $5$ in-context images cost
$\approx3,400$ tokens per query, nearly six times the cost of
zero-shot inference. At deployment, this token overhead translates
directly into latency and compute costs, and places a hard ceiling
on how many examples a practitioner can afford to include.

A natural alternative is to replace the images with text. Several recent lines of work have shown that class-level textual descriptions, attribute lists, GPT-generated category descriptions, or concept bottlenecks can substitute for or augment visual features in zero-shot classification with CLIP-style models \citep{menon2023visual, pratt2023cupl, yang2023labo}. We ask a different question, in a different regime: \emph{inside an MLLM, can per-example text, distilled offline from each individual training image, serve as the in-context signal in place of the image itself?}

The answer is not obvious in either direction. On one hand, text is cheap ($\approx 95$ tokens per identification rule vs. $\approx 550$ per image), and an MLLM that articulates its own visual reasoning may produce a context signal better aligned with how the model attends to images at inference. On the other hand, fine-grained discrimination: distinguishing an Abyssinian cat from a Bengal cat, or a 707-320 aircraft from an A320, is the regime where one expects visual grounding to be irreplaceable.

\begin{figure*}[t]
  \centering
  \includegraphics[width=1\textwidth]{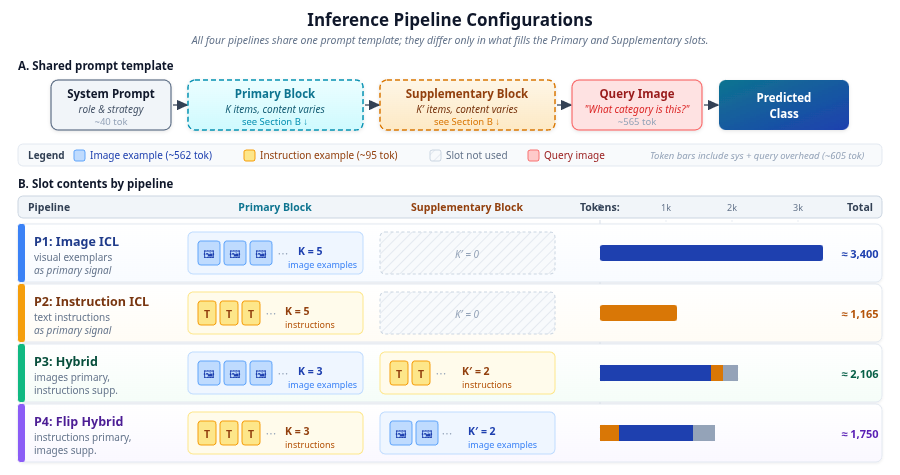}
  \vskip -4mm
  \caption{%
    \textbf{Inference pipeline overview.}
 The four context-construction strategies.
    \textbf{P1} (blue) places $K$ retrieved images in context.
    \textbf{P2} (orange) replaces them with $K$ text instructions,
    the most token-efficient configuration.
    \textbf{P3} (green) uses $K$ images as primary context and $K'$
    instructions as supplementary.
    \textbf{P4} (purple) reverses the role assignment: $K$ instructions
    are primary and $K'$ images are supplementary.
  }
  \label{fig:pipelines}
\end{figure*}

\textbf{Contributions.}
We propose \textit{instruction distillation}:\footnote{We use the term
\emph{distillation} in the sense of compressing the visual signal of an image into a compact textual form that retains the discriminative
content without pixel-level detail. It is not used in the standard knowledge-distillation sense, in which a student model learns from a teacher.} an offline procedure in which the MLLM itself is prompted, for each training image, to produce a structured identification rule capturing (i) general appearance cues for the depicted class, (ii) features that differentiate it from visually similar classes, and (iii) a common confusion point with a concrete visual test. At inference time, the top-$K$ training images are retrieved via CLIP similarity exactly as in image ICL, but their distilled instructions are placed in the model's context instead of images (Figure \ref{fig:instr}). These instructions are pregenerated; nothing is generated at inference time, which might add to latency. The key design choice is that instructions are generated per image, not per class: a single class can exhibit substantial intra-class visual variation, and grounding each instruction in a specific training instance preserves this diversity through retrieval, which class-level descriptions (e.g., \citet{menon2023visual}) cannot.

We evaluate two MLLM backbones across seven fine-grained benchmarks
under five inference configurations: zero-shot; image ICL;
instruction ICL; and two hybrid variants with opposite modality
orderings, holding the CLIP retrieval index fixed so any accuracy
difference is attributable to modality rather than retrieval. The
hybrid pipelines are motivated by two observations about retrieval.
In the first, retrieval quality degrades down the
ranked list: top neighbors are visually informative and best
supplied as images, while lower-ranked ones introduce noise
(irrelevant textures, viewpoint variation, near-miss classes) and
are better distilled into text. The second inverts this on the
grounds that in fine-grained settings, the top-ranked neighbors
often look nearly identical to the query yet belong to a different
class, so showing their images first can prime the model with a
misleading visual pattern. We instead present the top neighbors as
instructions, forcing the model to reason about discriminative
features before seeing any pixels. We find that:
\begin{itemize}[leftmargin=*, itemsep=-5pt, topsep=0pt]
    \item \textbf{Instruction ICL matches image ICL} at $K=1$ across all seven benchmarks and exceeds it on five of them, with a striking +57.9\% gain on EuroSAT~\citep{helber2019eurosatnoveldatasetdeep}, consistent with the hypothesis that, in the low-$K$ regime, a single retrieved image often presents an unrepresentative view of its class, whereas a distilled text rule encodes the general discriminative structure.
    \item \textbf{Instruction ICL reduces per-query input tokens} by 2.9× at $K=5$ and scales profitably to $K=25$ within the budget that image ICL exhausts at $K=5$, as seen in Table \ref{tab:token_cost}
    \item \textbf{Instruction ICL also cuts wall-clock inference latency}, and its one-time generation cost is recovered at scale: at $K=5$, mean per-query latency drops 3.27× (0.549s to 0.168s) and throughput improves 9.49×, exceeding the 2.9× token-count ratio because image tokens are more expensive to process per token than text tokens (Appendix~\ref{sec:appendix_efficiency}). 
    \item \textbf{Visual and textual ICL signals are complementary} under a fixed token budget: hybrid configurations consistently outperform either modality alone on benchmarks where both are individually useful, as seen in Figure \ref{fig:budget_line} (per-dataset bar-chart view in Appendix~\ref{sec:appendix_results}, Figure \ref{fig:budget_plot}). This shows that it is a compositional effect. Images give visual patterns to learn and see, while instructions give explicit rules and logic. When both of these are provided, the quality of context improves.
\end{itemize}
Together, these results reframe the role of in-context examples in MLLM visual classification: the question is not whether to use images or text, but what information the model needs from an example, and for many tasks, that information can be supplied at a fraction of the cost by letting the model articulate its own visual rules in advance.

\subsection{Related Work}
\label{sec:related}

\noindent\textbf{In-context learning.}
\citet{PLACEHOLDER_gpt3} showed that a frozen language model can
perform new tasks by conditioning on a handful of input--output
demonstrations at inference time, without parameter updates.
\citet{liu2022makesgoodexamples} and \citet{PLACEHOLDER_rubin2022learning}
showed that retrieving demonstrations similar to the query consistently
outperforms random selection. We adopt nearest-neighbor retrieval, but shift the question from \emph{which} demonstrations to
retrieve to \emph{in what modality} they should be presented.

\noindent\textbf{Multimodal in-context learning.} Flamingo extended ICL to MLLMs by
conditioning on interleaved image--text sequences
\citep{alayrac2022flamingo, PLACEHOLDER_mmicl}. Within this
setting, visual ICL substantially outperforms zero-shot prompting on
fine-grained classification, and visually similar retrieved examples
outperform random ones \citep{PLACEHOLDER_circle,jiang2024manyshot}.
These gains, however, come at a steep token cost, which motivates
our central question: whether the same neighbors can supply
useful ICL signal in textual form.

\noindent\textbf{Language descriptions of visual classes.}
A parallel line of work asks whether language can substitute for or
enrich visual representations.
\citet{menon2023visual} use GPT-generated descriptive
features as CLIP text prompts to improve zero-shot classification,
CuPL \citep{pratt2023cupl} extends this with LLM-generated
per-class appearance prompts, and LaBo \citep{yang2023labo}
constructs language-guided concept bottlenecks from GPT-3 sentences
for interpretable few-shot classification. All three methods share
two properties that distinguish them from ours. First, they produce a
\emph{single description per class}, collapsing intra-class visual
diversity into one canonical text. Second, they operate within the CLIP zero-shot framework: descriptions are encoded as text embeddings
and matched against image embeddings, with no in-context demonstrations and no MLLM in the loop.

\noindent\textbf{Our work in context.} We study a setting that no prior work does: an MLLM
performing in-context classification, where examples
are text instructions rather than images. Instructions are generated \emph{per training image}, preserving the
intra-class visual diversity that per-class descriptions collapse. Retrieval is held fixed against visual ICL, so accuracy differences
between modalities are attributable to demonstration content rather
than retrieval quality.

\section{Methodology}
\label{sec:method}

Let $\mathcal{D}_{\text{train}} = \{(x_i, y_i)\}_{i=1}^{N}$ denote a labeled
training set of image--label pairs, where $x_i$ is an image and
$y_i \in \mathcal{C}$ is a class label from a closed label set
$\mathcal{C} = \{c_1, \ldots, c_L\}$.
Given a test image $x_q$, the goal is to predict the correct label
$\hat{y}_q \in \mathcal{C}$ using a frozen multimodal large language
model $\mathcal{M}$, without any gradient updates to $\mathcal{M}$.

\noindent\textbf{In-context learning (ICL).}
In the ICL paradigm, $\mathcal{M}$ is conditioned on a prompt
$\mathcal{P}$ consisting of a sequence of \emph{demonstrations}
$\mathcal{E} = \{e_1, \ldots, e_K\}$ followed by the query $x_q$.
Each demonstration $e_k$ encodes evidence that the model should attend to
when classifying the query. The model produces the prediction as
\begin{equation}\label{eq:icl}
  \hat{y}_q = \arg\max\nolimits_{c \in \mathcal{C}}\;
    p_{\mathcal{M}}\!\left(c \mid \mathcal{P}(\mathcal{E}, x_q)\right).
\end{equation}
In standard \emph{visual} ICL, each demonstration is an image-label pair:
$e_k = (x_k, y_k)$, where $x_k$ is a training image and $y_k$ its label.
The central question of this work
is whether visual demonstrations can be replaced by textual
demonstrations without sacrificing classification accuracy.


\subsection{Instruction Distillation}
\label{sec:distillation}

Standard image ICL places the burden of visual reasoning entirely on
the model at inference time: the model must extract discriminative
features from each example image, relate them to the query, and
arrive at a prediction in a single forward pass. We hypothesize that
much of this work can be offloaded to an offline step in which the
model articulates, in natural language, the visual rules that
characterize a class. Examples of this are given in Appendix \ref{sec:appendix_instructions}. The resulting instructions can then serve as
context at inference time, at a fraction of the token cost of the
underlying image.

\noindent\textbf{Instruction generation.}
For each training image $x_i$ with label $y_i$, we prompt the same
MLLM $\mathcal{M}$ to generate a structured identification
instruction $t_i$ consisting of three rules:
\begin{enumerate}[leftmargin=*, itemsep=-1pt, topsep=0pt]
  \item \textbf{General visual cues:} shape, color, texture, or
    structural properties characteristic of the class.
  \item \textbf{Differentiating features:} visual characteristics
    that distinguish this class from its nearest neighbors in the
    embedding space.
  \item \textbf{Common confusion point:} a specific class frequently
    confused with this one, and a concrete visual test that resolves
    the ambiguity.
\end{enumerate}
This three-part structure is designed to surface the contrasts most useful for fine-grained discrimination: what the class looks like, how it differs from neighbors, and where it is mistakenly identified rather than producing a generic description. The prompt template is domain-adapted across datasets. The full
template is given in Appendix~\ref{sec:appendix_prompts} for Food-101 dataset~\citep{10.1007/978-3-319-10599-4_29}.

\noindent\textbf{Per-image, not per-class.}
Instructions are generated independently for each training image
rather than once per class, a deliberate departure from prior work
on language descriptions for visual classification
\citep{menon2023visual, pratt2023cupl, yang2023labo}, which
collapses each class into a single canonical text. Two
considerations motivate this choice. First, a class can exhibit
substantial intra-class visual variation: instances of
\textit{pizza} differ in toppings, crust, cut, and lighting, and
instructions grounded in different images capture this diversity.
Second, grounding each instruction in a specific visual instance
produces more concrete and actionable rules than a class-level
description generated without visual input. At retrieval time,
selecting instructions from the most similar training images
provides query-relevant guidance that a per-class description
cannot.

\noindent\textbf{Instruction store.}
Generated instructions are stored in a dictionary keyed by training image: $\mathcal{T} = \{x_i \mapsto (y_i, t_i)\}$. It is a one-time offline effort with a cost proportional to the training set size. 
\begin{table}[t]
  \centering
  \small
  \begin{tabular}{lrr}
    \toprule
    \textbf{Configuration} & \textbf{Tokens} \\
    \midrule
    P0: Zero-Shot           & $\approx$574  \\
    P1(a): Image ICL ($K{=}5$)         & $\approx$3,400 \\
    P1(b): Image-compressed ICL ($K{=}5$) & $\approx$1940 \\
    P2: Instruction ICL ($K{=}5$)      & $\approx$1,165 \\
    P3: Hybrid ($K{=}3$, $K'{=}2$)    & $\approx$2,106 \\
    P4: Flip Hybrid ($K{=}3$, $K'{=}2$)  & $\approx$1,950 \\
    \bottomrule
  \end{tabular}
  \caption{\textbf{Token Cost.} Approximate input token cost for each inference
    configuration at $K{=}5$.}
  \label{tab:token_cost}
\end{table}

\subsection{Inference Pipelines}
\label{sec:pipelines}

We define five inference configurations that systematically vary which
modality: image, text instruction, or both is used as in-context
evidence. All five share the same CLIP retrieval backbone
(Appendix~\ref{sec:retrieval}) and the same frozen MLLM $\mathcal{M}$. Figure~\ref{fig:pipelines} illustrates the five configurations and 
Table~\ref{tab:token_cost} summarizes the token cost of each
configuration.

\noindent\textbf{P0: Zero-Shot.}
P0 provides no demonstrations: the prompt contains only the query
image and a brief classification instruction. This establishes the
baseline capability of $\mathcal{M}$.

\noindent\textbf{P1: Image ICL.}
P1 is the standard retrieval-augmented visual ICL baseline: the
top-$K$ retrieved training images are placed in context as
demonstrations, each paired with its ground-truth label. The prompt
structure is:
\[
  \mathcal{P}_{\text{img}} = \bigl[e_1^{\text{img}}, \ldots,
    e_K^{\text{img}},\; x_q\bigr],
  \quad e_k^{\text{img}} = (x_k, y_k).
\]
We test two variants of this pipeline: P1(a) uses the true full resolution image as context, and correspondingly has a higher token consumption. P1(b) compresses all images to a size of about 224x152, while retaining 75\% of original quality. This occupies $\approx258$ tokens, about 54\% reduction.

\begin{figure*}[h]
  \centering
  \includegraphics[width=1\textwidth]{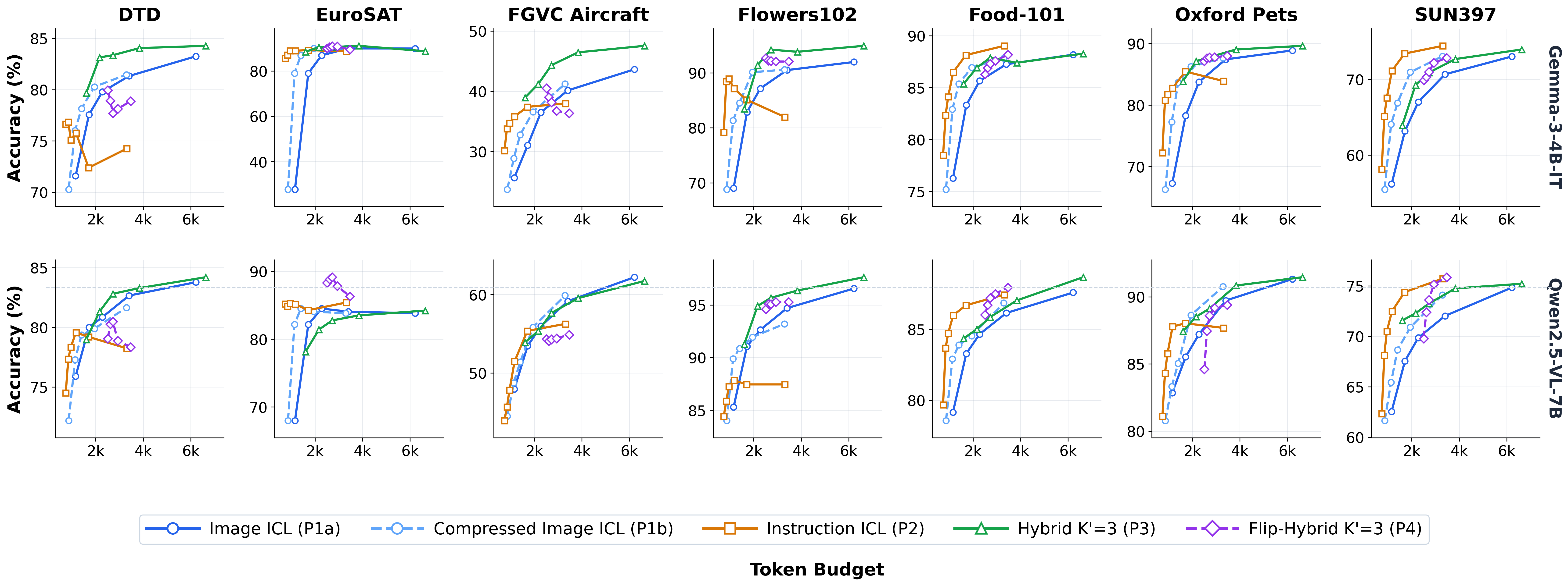}
  \vskip -4mm
  \caption{
    \textbf{Accuracy versus token budget.} Mean per-class accuracy (\%) plotted against input tokens for P1(a) (solid blue), P1(b) (dashed blue), P2 (orange), P3 with $K' = 3$ (green), and P4 with $K' = 3 $ (purple). Top row: Gemma-3-4B-IT. Bottom row: Qwen2.5-VL-7B-Instruct. Each marker corresponds to one $(K)$ or $(K, K')$ configuration; markers are connected in order of increasing token cost.
}
  \label{fig:budget_line}
\end{figure*}

\noindent\textbf{P2: Instruction ICL.}
P2 is our proposed replacement for image demonstrations: the $K$
retrieved training images are used only to look up their distilled
instructions, which are placed in context in lieu of the images
themselves. The query image is still provided. Since the retrieval
index is identical to P1, any accuracy difference between the two is
attributable to modality rather than to retrieval quality. The prompt
structure is:
\[
  \mathcal{P}_{\text{inst}} = \bigl[e_1^{\text{inst}}, \ldots,
    e_K^{\text{inst}},\; x_q\bigr],
  \quad e_k^{\text{inst}} = (y_k, t_k).
\]
P2 is the most token-efficient configuration in our
suite, evident in Table \ref{tab:token_cost}.

\noindent\textbf{P3: Hybrid ICL (image-primary).}
P3 combines both modalities, exploiting the fact that retrieval
quality degrades down the ranked list. The top-$K$ neighbors, which
carry the strongest visual similarity to the query, are presented as
images to supply clear perceptual evidence. The next $K'$ neighbors
(non-overlapping with the top-$K$) are converted to text instructions:
at this depth, images begin to introduce irrelevant textures,
viewpoint variation, and near-miss classes, whereas the distilled
instructions retain the class-discriminative content while discarding
this visual noise. The system-prompt frames images as primary
evidence and instructions as supplementary reasoning guidance.

\noindent\textbf{P4: Flip Hybrid ICL (instruction-primary).}
P4 reverses the modality roles of P3. In fine-grained settings, the
top-ranked neighbors often look nearly identical to the query, yet
belong to a different class, so presenting their images first can
prime the model with a misleading visual pattern. P4 instead
presents the top-$K$ neighbors as instructions, forcing the model to
reason about discriminative class features before seeing any pixels;
the next $K'$ neighbors are then provided as
images to supply visual grounding for the prediction. The system
prompt frames instructions as primary guidance and images as
corroborating evidence.



\section{Experiments}
\label{sec:experiments}

\subsection{Experimental Setup}
\label{sec:setup}

\noindent\textbf{Datasets.}
We evaluate on seven fine-grained visual classification benchmarks spanning
diverse visual domains: Food-101~\citep{10.1007/978-3-319-10599-4_29}
(101 food categories, 25,250 test images), 
Oxford-IIIT Pets~\citep{parkhi12a} (37 pet breeds,
3,699 test images), DTD ~\citep{cimpoi2013describingtextureswild} (47 texture
categories, 1,692 test images), EuroSAT~\citep{helber2019eurosatnoveldatasetdeep}
(10 satellite land-use classes, 8,100 test images), Flowers102 ~\citep{flower_Nilsback08} (102 flower classes, 2,463 test images), SUN397~\citep{sun_inproceedings} (397 scene categories, 19,850 test images) and FGVC~\citep{maji2013finegrainedvisualclassificationaircraft} (100 airplane classes, 3,333 test images). All datasets are used with their standard train/test splits, as in the experiments of \citet{PLACEHOLDER_circle}.
Full dataset statistics are provided in
App.~\ref{sec:appendix_datasets}.

\noindent\textbf{Models.} We evaluate two publicly available MLLMs:
Gemma-3-4B-IT and Qwen2.5-VL-7B-Instruct.
In all pipelines, in-context examples are selected by nearest-neighbor retrieval over CLIP ViT-B/32 embeddings. Implementation details are in App.~\ref{sec:appendix_impl}, context selection in App.~\ref{sec:retrieval}.

\noindent\textbf{Inference Configurations.}
We evaluate five inference configurations, described in detail in Section \ref{sec:method}. For the unimodal pipelines (P1, P2) we evaluate $K \in $\{1, 2, 3, 5, 10\}, with P2 additionally evaluated at $K = 25$ to test scaling behavior within the token budget that P1 exhausts at $K = 5$. For the two hybrid pipelines (P3, P4) we evaluate a grid of $(K, K') \in $\{1, 2, 3, 5, 10\} × \{1, 2, 3\}. Per-pipeline token-cost formulas are derived in App.~\ref{sec:appendix_tokens}.

\noindent\textbf{Evaluation.}
We report \textit{mean per-class accuracy} (macro-averaged top-1
accuracy), which is robust to class imbalance. To enable
cross-pipeline comparison, we define three per-query token budget
regimes: \textbf{Low} ($\sim$1.5k tokens), \textbf{Mid} ($\sim$3.3k
tokens), and \textbf{High} ($\sim$6.3k tokens). For each pipeline,
the $(K, K')$ configuration closest to each budget is reported in
Table~\ref{tab:budget_main}, with the exact token cost of each
selected configuration. P2 and P4 are dominated by text instructions
and do not reach High budget in any of our $(K, K')$
configurations, so the High-budget row contains entries only for P1
and P3 (see Figure~\ref{fig:avg_budget}; per-dataset results in App.~\ref{sec:appendix_results}, Figure~\ref{fig:budget_plot}).

\begin{table}[t]
\centering
\footnotesize
\resizebox{\linewidth}{!}{
\begin{tabular}{lllllll}
\toprule
\textbf{K} & \textbf{Per-class} & \textbf{Caption} & \textbf{Single rule} & \textbf{Paragraph} & \textbf{2-rule} & \textbf{3-rule} \\
\midrule
1  & 0.6609 & 0.6178 & 0.6140 & 0.6361 & \textbf{0.7286} & 0.7224 \\
5  & 0.6972 & 0.7893 & 0.7856 & 0.7830 & 0.8304 & \textbf{0.8276} \\
10 & 0.6565 & 0.7860 & 0.7920 & 0.8010 & 0.8215 & \textbf{0.8545} \\
25 & 0.6672 & 0.7544 & 0.7678 & 0.7711 & 0.8124 & \textbf{0.8390} \\
\bottomrule
\end{tabular}
}
\resizebox{\linewidth}{!}{
\begin{tabular}{lllllll}
\toprule
\textbf{K} & \textbf{Per-class} & \textbf{Caption} & \textbf{Single rule} & \textbf{Paragraph} & \textbf{2-rule} & \textbf{3-rule} \\
\midrule
1  & 0.2556 & 0.1992 & 0.2044 & 0.2172 & \textbf{0.3254} & 0.3015 \\
5  & 0.2877 & 0.3233 & 0.3343 & 0.3306 & 0.3425 & \textbf{0.3582} \\
10 & 0.2633 & 0.33 & 0.3483 & 0.3373 & 0.3546 & \textbf{0.3741} \\
25 & 0.2731 & 0.3273 & 0.3481 & 0.3555 & 0.3365& \textbf{0.3798} \\
\bottomrule
\end{tabular}
}
\vskip -2mm
\caption{\textbf{Ablation study with PETS and FGVC.} Comparison of three
basic instruction formats: caption, paragraph, and a single rule with our 3-rule format, which outperforms all alternatives across $K$, indicating that both structured decomposition and the inclusion of general and
error-avoidance cues contribute to performance.}
\label{tab:appendix_ablation}
\end{table}

\subsection{Main Results}
\label{sec:main_results}

\definecolor{budgetlow}{HTML}{F4EBE1}
\definecolor{budgetmid}{HTML}{E6EEF6}
\definecolor{budgethigh}{HTML}{E6F2EA}
\definecolor{modelband}{HTML}{D5DEE7}

\begin{table*}[t]
\centering\footnotesize
\renewcommand{\arraystretch}{1.15}
\setlength{\tabcolsep}{5pt}
\begin{tabular}{lrccccccc|c}
\toprule
\textbf{Pipeline} & \textbf{Tokens} & \textbf{PETS} & \textbf{FOOD} & \textbf{DTD} & \textbf{ESAT} & \textbf{FGVC} & \textbf{FLWR} & \textbf{S397} & \textbf{Avg.} \\
\midrule
\rowcolor{modelband}\multicolumn{10}{l}{\textbf{\textsc{Gemma-3-4B-IT}}} \\
\rowcolor{budgetlow}\multicolumn{10}{l}{\textit{Low budget ($\sim$1.5k tokens)}} \\
P1(a) ($K{=}1$)           & 1{,}152 & 67.3          & 76.3          & 71.6          & 27.8          & 25.7          & 69.0          & 56.2          & 56.3 \\
P1(b) ($K{=}3$)           & 1{,}402 & 83.7          & 85.4          & 78.2          & 86.9          & 32.8          & 84.5          & 66.9          & 74.0 \\
P2 ($K{=}10$)             & 1{,}700 & \textbf{85.5} & \textbf{88.1} & 72.4          & 89.2          & 37.4          & 85.1          & \textbf{73.4} & 75.9 \\
P3 ($K{=}1, K'{=}3$)      & 1{,}493 & 83.9          & 85.4          & \textbf{79.7} & 88.5          & \textbf{38.9} & 83.4          & 63.9          & 74.8 \\
P4 ($K{=}3, K'{=}1$)      & 1{,}600 & 85.4          & 86.2          & 77.0          & \textbf{90.0} & 34.0          & \textbf{89.4} & 69.7          & \textbf{75.9} \\
\rowcolor{budgetmid}\multicolumn{10}{l}{\textit{Mid budget ($\sim$3.3k tokens)}} \\
P1(a) ($K{=}5$)           & 3{,}400 & 87.4          & 87.3          & 81.3          & 90.0          & 40.1          & 90.5          & 70.7          & 78.2 \\
P1(b) ($K{=}10$)          & 3{,}292 & 87.6          & 87.8          & 81.5          & 89.9          & 41.2          & 90.6          & 73.0          & 78.8 \\
P2 ($K{=}25$)             & 3{,}305 & 83.9          & \textbf{89.0} & 74.3          & 88.7          & 38.0          & 82.0          & \textbf{74.4} & 75.8 \\
P3 ($K{=}5, K'{=}3$)      & 3{,}729 & \textbf{89.0} & 87.4            & \textbf{84.0} & \textbf{91.1} & \textbf{46.5} & \textbf{93.8} & 72.7          & \textbf{80.6} \\
P4 ($K{=}10, K'{=}3$)     & 3{,}360 & 88.0          & 88.2          & 78.9          & 89.5          & 36.4          & 92.1          & 72.8          & 78.0 \\
\rowcolor{budgethigh}\multicolumn{10}{l}{\textit{High budget ($\sim$6.3k tokens)}} \\
P1(a) ($K{=}10$)          & 6{,}210 & 88.9          & 88.2 & 83.2 & \textbf{89.9} & 43.7          & 92.0          & 73.0          & 79.8 \\
P3 ($K{=}10, K'{=}3$)     & 6{,}631 & \textbf{89.6} & \textbf{88.3}            & \textbf{84.3}            & 88.7          & \textbf{47.6} & \textbf{94.9} & \textbf{73.9} & \textbf{81.0} \\
\midrule
\rowcolor{modelband}\multicolumn{10}{l}{\textbf{\textsc{Qwen2.5-VL-7B-Instruct}}} \\
\rowcolor{budgetlow}\multicolumn{10}{l}{\textit{Low budget ($\sim$1.5k tokens)}} \\
P1(a) ($K{=}1$)           & 1{,}152 & 82.9          & 79.2          & 75.9          & 68.0          & 47.9          & 85.3          &  62.3           & 71.6 \\
P1(b) ($K{=}3$)           & 1{,}402 & 85.0          & 83.9          & \textbf{79.4} & 84.5          & 51.4          & 90.9          & 68.7            & 77.7 \\
P2 ($K{=}10$)             & 1{,}700 & \textbf{88.0} & \textbf{86.7} & 79.2          & 84.3          & \textbf{55.4} & 87.5          & \textbf{74.4 }           & \textbf{79.4} \\
P3 ($K{=}1, K'{=}3$)      & 1{,}493 & 87.4          & 84.4          & 79.0          & 78.2          & 53.9          & 91.3          & 71.6            & 78.0 \\
P4 ($K{=}3, K'{=}1$)      & 1{,}600 & 81.9          & 84.7          & 78.6          & \textbf{87.0} & 48.4          & \textbf{93.3} & 71.9            & 78.0 \\
\rowcolor{budgetmid}\multicolumn{10}{l}{\textit{Mid budget ($\sim$3.3k tokens)}} \\
P1(a) ($K{=}5$)           & 3{,}400 & 89.7          & 86.2          & 82.7          & 84.1          & 59.2          & 94.7          & 72.0            & 81.2 \\
P1(b) ($K{=}10$)          & 3{,}292 & \textbf{90.8} & 86.9          & 81.7          & 83.8          & \textbf{59.9} & 93.2          & 74.1            & 81.5 \\
P2 ($K{=}25$)             & 3{,}305 & 87.7          & 87.4          & 78.2          & 85.4          & 56.3          & 87.5          & 75.7            & 79.7 \\
P3 ($K{=}5, K'{=}3$)      & 3{,}729 & \textbf{90.8} & 87.0          & \textbf{83.3} & 83.5          & 59.6          & \textbf{96.4} & 74.8            & \textbf{82.2} \\
P4 ($K{=}10, K'{=}3$)     & 3{,}360 & 89.4          & \textbf{88.0} & 78.4          & \textbf{86.3} & 54.9          & 95.3          & \textbf{75.9}            & 81.2 \\
\rowcolor{budgethigh}\multicolumn{10}{l}{\textit{High budget ($\sim$6.3k tokens)}} \\
P1(a) ($K{=}10$)          & 6{,}210 & 91.3          & 87.6          & 83.8          & 83.8          & \textbf{62.2} & 96.6          & 74.8            & 82.9 \\
P3 ($K{=}10, K'{=}3$)     & 6{,}631 & \textbf{91.5} & \textbf{88.7} & \textbf{84.2} & \textbf{84.2} & 61.7          & \textbf{97.6} & \textbf{75.2 }           & \textbf{83.3} \\
\bottomrule
\end{tabular}
\vskip -2mm
\caption{\textbf{Pipeline comparison across token budgets and
  seven fine-grained benchmarks.} Mean per-class accuracy (\%) for P1(a), P1(b), P2, P3 and P4 on Gemma-3-4B-IT (top) and Qwen2.5-VL-7B-Instruct
  (bottom), evaluated at three token budgets:
  \emph{low} ($\sim$1.5k), \emph{mid} ($\sim$3.3k), and \emph{high}
  ($\sim$6.3k). The \emph{Tokens} column reports the actual input cost
  of the chosen $(K, K')$ configuration. Best per dataset within each
  budget block is shown in \textbf{bold}.
 }
\label{tab:budget_main}
\end{table*}


Table \ref{tab:budget_main} reports mean per-class accuracy for each of the three token budgets, for both backbones. Figure \ref{fig:budget_line} disaggregates the same results by dataset (bar-chart view in Appendix~\ref{sec:appendix_results}, Figure \ref{fig:budget_plot}), and Figure \ref{fig:avg_budget} summarizes the cross-dataset averages.
Two findings characterize the table. First, on aggregate, the \textbf{image-primary hybrid P3 is the best pipeline at every budget} on both backbones, averaging 74.8–81.0\% on Gemma and 78.0–83.3\% on Qwen across Low, Mid, and High budgets. Second, at the Low budget, the three instruction-using pipelines (P2, P3, P4) \textbf{all substantially exceed pure image ICL}: on Gemma, the gap is  $\approx$\textbf{+20pp} (56.3\% for P1 vs. 74.8–75.9\% for the others), and on Qwen, approximately \textbf{+7pp} (71.6\% vs. 78.0–79.4\%). Section \ref{sec:analysis} examines per-dataset distribution of these gaps, mechanisms that produce them, and conditions under which they reverse.

\subsection{Ablation Study}
We conduct three ablations to validate our instruction generation
pipeline with PETS \citep{parkhi12a} and FGVC \citep{maji2013finegrainedvisualclassificationaircraft}. See Table \ref{tab:appendix_ablation} for combined results. \emph{First}, we compare against (i) a single-sentence caption,
(ii) an unstructured 2--3 sentence descriptive paragraph,  (iii)
a single discriminative rule focused only on inter-class separation, and (iv) using a 2-rule format (general + discriminative) dropping the confusion point rule.
\textbf{None outperforms our 3-rule baseline.}
We attribute this to the caption and paragraph lacking an explicit
discriminative objective, producing generic descriptions that fail
to distinguish visually similar images, while the single
discriminative rule omits the complementary general and
error-avoidance cues, and the two-rule format also fails because the number of examples increases in the context; the importance of the "common confusion point" also increases to steer the model in the right direction. \emph{Second}, we replace per-image instructions with
a single canonical per-class instruction. The observed weaker performance
indicates that \textbf{class-level information alone is insufficient}: it
fails to capture intra-class variation, motivating the per-image
design (App.~\ref{sec:appendix_ablation}).
\emph{Third}, we compare per-image instruction ICL against a label-only control, in which instructions are generated from the identical prompt (App.~\ref{sec:appendix_prompts}) but with the training image removed, to test whether the generated rules are driven by label-conditioned class priors rather than image-specific visual evidence. Per-image instructions never underperform the label-only control at any tested $K$ on either dataset and are substantially less repetitive: the exact-duplicate ratio falls from 0.653 for label-only instructions to 0.017 for per-image instructions. See App.~\ref{sec:appendix_ablation} for the full lexical and semantic diversity analysis.

\section{Analysis}
\label{sec:analysis}
We organize this section around the question: at a fixed per-query token budget, which inference pipeline maximizes accuracy, and why? Numbers are from Table~\ref{tab:budget_main}; per-dataset results from App.~\ref{sec:appendix_results}.

\subsection{Budget-Bucketed Comparison}
\begin{figure}
    \centering
    \includegraphics[width=1\linewidth]{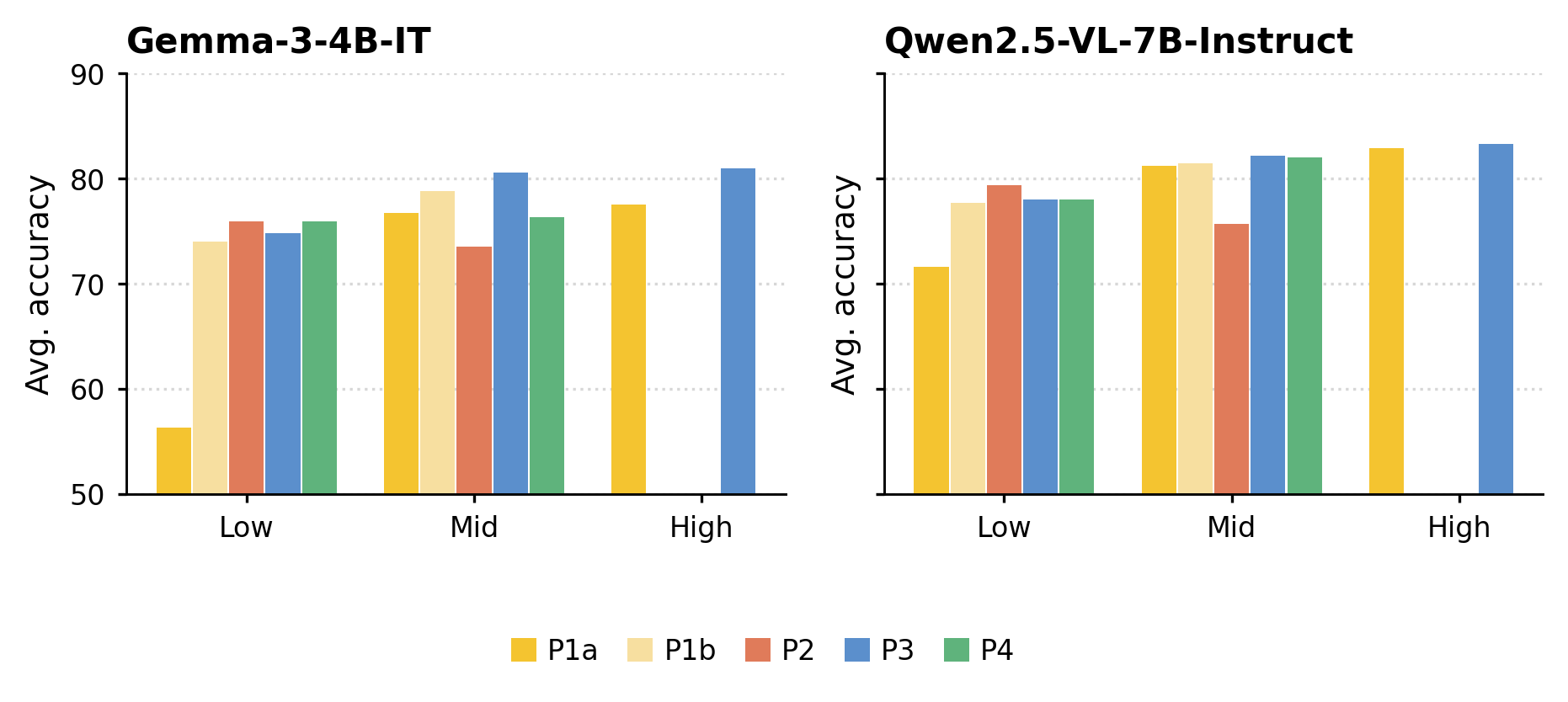}
    \vskip -4mm
    \caption{\textbf{Average accuracy by pipeline and token budget.} Mean per-class accuracy averaged across seven fine-grained benchmarks under three token budgets, for P1(a), P1(b), P2, P3, and P4. Instruction-based pipelines (P1(b), P2, P3, P4) substantially outperform full-resolution image ICL (P1(a)) at the Low budget.}
    \label{fig:avg_budget}
\end{figure}

Results in Table~\ref{tab:budget_main} and
Figure~\ref{fig:avg_budget} reveal three patterns. First, at the Low
budget ($\sim$1.5k tokens) on Gemma-3-4B-IT, P1(a) averages $56.3\%$
across the seven benchmarks, whereas P2, P3, and P4 average
$75.9\%$, $74.8\%$, and $75.9\%$. The gap of roughly \textbf{+20 pp}
between P1(a) and any instruction-using pipeline is the largest single
effect in our experiments. On Qwen2.5-VL-7B-Instruct, the same
ordering holds at a smaller magnitude of \textbf{+7 pp}
($71.6\%$ vs. $78.0$--$79.4\%$). Second, as the budget grows, P1(a)
closes most of the gap by accumulating additional in-context images
($78.2\%$ at Mid, $79.8\%$ at High on Gemma), but \textbf{P3 retains
a consistent lead at every budget evaluated}, reaching $81.0\%$ at
High budget on Gemma against P1(a)'s $79.8\%$. On Qwen, the
corresponding figures are $83.3\%$ vs. $82.9\%$. Third,
\textbf{neither P1(a) nor P1(b) is ever the highest-scoring pipeline
at any budget on either backbone}. The implication is  direct: across
all budgets, P3 dominates P1(a), and at the Low budget, all
three instruction-using pipelines substantially exceed P1(a) on both
MLLMs.

\subsection{EuroSAT as a Limiting Case}
EuroSAT~\citep{helber2019eurosatnoveldatasetdeep} is the most
informative dataset for understanding why instructions
help. At Low budget on Gemma, P1(a) reaches $27.8\%$ while P2
reaches $89.2\%$ - a gap of \textbf{61.4 points}. The gap on Qwen
is \textbf{16.3 points}, and P2 on Gemma approaches the asymptotic accuracy of the dataset.
We read this as a failure mode of the low-budget image ICL
rather than a property of either backbone. Satellite tiles for a
given land-use class vary in color grading, season, cloud cover, so a retrieved neighbor at low budget often shares
low-level spectral statistics with the query rather than
class-discriminative content. A distilled instruction, by contrast,
names the class and lists its diagnostic features in stable
language. The persisting gap on Qwen supports this: when intra-class variance is high relative to
inter-class variance, image-only context at a tight budget is a poor
source of evidence regardless of the consuming model.

\subsection{Hybrid Configurations Dominate}
P3 is the most consistent winner across our experiments. At mid-budget on Gemma, \textbf{P3 matches or exceeds P1(a) on all seven datasets}, with the largest margins on FGVC (\textbf{+6.4}), FLWR (\textbf{+3.3}), DTD (\textbf{+2.7}), and S397 (\textbf{+2.0}). On Qwen at the mid-budget, P3 exceeds P1(a) on six of the seven datasets, the sole exception being ESAT, where P1(a) leads by 0.6 points. At high-budget, P3 again exceeds P1(a) across all datasets (except ESAT for Gemma and FGVC for Qwen).
Complementarity is most informative in datasets where the unimodal baselines are individually weak. On FGVC at mid-budget, neither P1(a) (40.1\%) nor P2 (38.0\%) is competitive, yet P3 reaches 46.5\%. On FLWR, P2 plateaus at 82.0\% and P1(a) reaches 90.5\%, while P3 reaches 93.8\%. In both cases, the gain from the hybrid exceeds contributions of either modality in isolation, indicating that images and instructions correct disjoint classes of failures.
A natural interpretation is that the two modalities compose. When MLLM
is fed images, each successive one is highly correlated
with the previous. The model's understanding saturates quickly;
switching modality midway introduces a structurally different
signal, even with fewer examples per modality, boosting
performance. Images can mislead when retrieved neighbors share
surface texture with the query but belong to a confusable class;
instructions can be uninformative when the discriminative cue is
perceptual and resists verbalization. In hybrid,
the model has two semi-independent signals to cross-validate.

\subsection{Instruction Quality}
We used an LLM judge (Gemma4-31b-it) to quantitatively evaluate instruction quality across 100 random images per dataset on three criteria: groundedness, genericness, and hallucination (each on a 1-5 scale). Mean groundedness is 4.57, and hallucination is 3.87, indicating that generated instructions largely reference real, visible features; genericness is lower (3.20), reflecting moderate label-prior tendency. ESAT is a clear outlier, with the lowest groundedness (3.30) and hallucination (2.63) scores, consistent with the abstract, low-texture nature of satellite imagery. Methodology, per-dataset scores, and human validation are in App.~\ref{sec:appendix_quality}.

\subsection{Boundary Conditions and Failure Modes}
P2 underperforms P1(a) on datasets where discriminative cues are predominantly perceptual. At the mid-budget on Qwen, P2 trails P1(a) by 7.2 points on FLWR (87.5\% vs. 94.7\%) and by 2.9 points on FGVC (56.3\% vs. 59.2\%). Flowers and aircraft variants are distinguished by petal geometry, fuselage proportions, and engine configurations that are difficult to verbalize precisely but readily perceived. Crucially, the image-primary hybrid recovers full performance on both (96.4\% on FLWR, 59.6\% on FGVC at the Mid budget), indicating that the limitation is specific to the pure-instruction variant rather than the broader method.
On DTD, P2's advantage over P1(a) reverses as the budget increases on Gemma. At the Low budget, P2 reaches 72.4\%, exceeding P1(a) (71.6\%); by the Mid budget, P2 has only inched up to 74.3\%, while P1(a) climbs to 81.3\% and overtakes it. We attribute this to redundancy: as the budget admits more retrieved instructions, those instructions become increasingly similar in content and dilute rather than reinforce the discriminative signal, whereas additional retrieved images continue to add perceptual information.

\paragraph{Conclusion}
We introduced \textit{instruction distillation}, a procedure in
which a MLLM uses a compact,
structured identification rule that replaces the image in context at inference time. Across seven benchmarks and
two MLLMs, instruction-only ICL matches or exceeds image
ICL in the low-shot regime at a fraction of the token cost, and
scales to context sizes image ICL cannot reach within the same
budget. Hybrid configurations exceed both unimodal baselines under
matched token budgets, showing that visual and textual ICL signals
compose rather than substitute. These results show that images can
be distilled into text without sacrificing, and often improving,
classification accuracy. Extensions include relaxing the
closed-set restriction to multi-label and open-vocabulary
classification, testing cross-model transferability of distilled
instructions, and exploring richer instructions.

\newpage

\section*{Limitations}
Several factors bound the scope of our findings. First, instruction quality is upper-bounded by the visual understanding of the generator MLLM: a weaker generator may produce generic or hallucinated rules that may mislead the inference model. Second, instruction distillation incurs a one-time offline cost proportional to the size of the training set; while this is amortized across all subsequent test queries, the upfront generation budget can be substantial for very large training corpora. 




\bibliography{custom}

\appendix
\input{appendix}

\end{document}

%% file: appendix.tex
\clearpage
\twocolumn[{%
    \centering
    \vspace{0.5em}
    {\LARGE \textbf{Instruction Distillation: Text Instructions as Visual Examples}}\\[0.5em]
    {\large Appendix}
    \vspace{1em}
}]
This section provides additional technical
and implementation details, alongside extended experimental results and further analyses that complement the main
paper. Specifically, the material is organized as follows:
\begin{itemize}[noitemsep]
\item Appendix \ref{sec:appendix_datasets} provides details regarding the datasets.
\item  Appendix \ref{sec:appendix_impl} lists the exact implementation details for various experiments performed.
\item  Appendix \ref{sec:retrieval} provides details about how examples are retrieved and rationale behind it
\item  Appendix \ref{sec:appendix_results} lists in details results for all inference pipelines at various K and K' values.

\item  Appendix \ref{sec:appendix_prompts} provides example prompt which we used, specifically for the Food-101 dataset.
\item  Appendix \ref{sec:appendix_tokens} details the token calculation used in order to evaluate all inference pipelines.
\item Appendix \ref{sec:appendix_quality} reports the LLM-judge evaluation of instruction quality (groundedness, genericness, hallucination) summarized in Section \ref{sec:analysis}.
\item Appendix \ref{sec:appendix_ablation} details the ablation study results and conclusions as discussed on Section \ref{sec:analysis}.
\item Appendix \ref{sec:appendix_instructions} illustrates various images from various datasets and instructions generated.
\end{itemize}
\section{Datasets}
\label{sec:appendix_datasets}

We evaluate our approach on the following fine-grained visual
classification benchmarks. All datasets are used in their standard
train/test splits without any modification. We summarize the evaluation datasets in Table \ref{tab:datasets}

\begin{table}[H]
\centering\small
\begin{tabular}{llrrr}
\toprule
\textbf{Abbr.} & \textbf{Dataset} & \textbf{Images} & \textbf{Classes} \\
\midrule
FOOD   & Food-101         & 25,250  & 101 \\
DTD    & DTD        & 1,692   & 47 \\
FGVC   & FGVC Aircraft       & 3,333   & 100 \\
PETS   & Oxford-IIIT Pets    & 3,699   & 37 \\
ESAT & EuroSAT         & 8,100  & 10 \\
FLWR & Flowers102 & 2,463 & 102 \\
S397 & SUN397 & 19,850 & 397 \\
\bottomrule
\end{tabular}
\caption{\textbf{Dataset details.} Summary details of the datasets used in
our experiments.}
\label{tab:datasets}
\end{table}

\section{Implementation Details}
\label{sec:appendix_impl}

\paragraph{Hardware.}
All experiments are conducted on NVIDIA A100 GPUs with 80\,GB HBM2e
memory. Models up to 7B parameters (Gemma-3-4B, Phi-3.5-Vision,
Qwen2.5-VL-7B) are served on a single A100. Evaluation time ranges from a few minutes 
for small datasets like DTD and FLWR to 8-10 hours for the
largest datasets like FOOD and S397 (See Table \ref{tab:datasets} for details on their sizes). Instruction generation was relatively a faster step and took few minutes to an hour (maximum). For inference, a temperature of 0.2 is set for all pipelines and maximum output tokens is set at 150 for inference and 512 for offline instruction generation.




\section{Retrieval-Augmented Context Selection}
\label{sec:retrieval}

Rather than selecting demonstrations randomly or using a fixed support set, we retrieve the $K$ training examples most visually similar to the query
image. This design is motivated by prior work showing that retrieved
demonstrations that share the query's visual characteristics are more
informative than random examples for few-shot visual classification.

\paragraph{CLIP embeddings.}
We embed all training images using a pre-trained CLIP
ViT-B/32 encoder with OpenAI weights. All embeddings are
$\ ell_2$-normalized to unit length, so cosine similarity reduces to a
dot product. Let $\phi(x) \in \mathbb{R}^{512}$ denote the normalised
CLIP embedding of image $x$. The similarity between the training image $x_i$
and query $x_q$ is given by
\begin{equation}
  s(x_i, x_q) = \phi(x_i)^\top \phi(x_q).
  \label{eq:sim}
\end{equation}


\paragraph{Why visual retrieval for instruction ICL?}
A key design choice is to use visual (CLIP) similarity to retrieve
instructions, rather than a text-based retrieval over instruction content.
Instructions are anchored to specific training images; retrieving
instructions from visually similar images increases the likelihood that the
retrieved rules reference appearance properties that are actually visible
in the query image (e.g., the same plating style, the same
textural variation). Class-level retrieval would miss this intra-class
visual diversity.

\section{Full Results}
\label{sec:appendix_results}

This section provides complete per-dataset, per-model result tables
supplementing the aggregated results in the main paper.
Each table reports mean per-class accuracy (\%).

\begin{figure*}[t]
  \centering
  \includegraphics[width=1\textwidth]{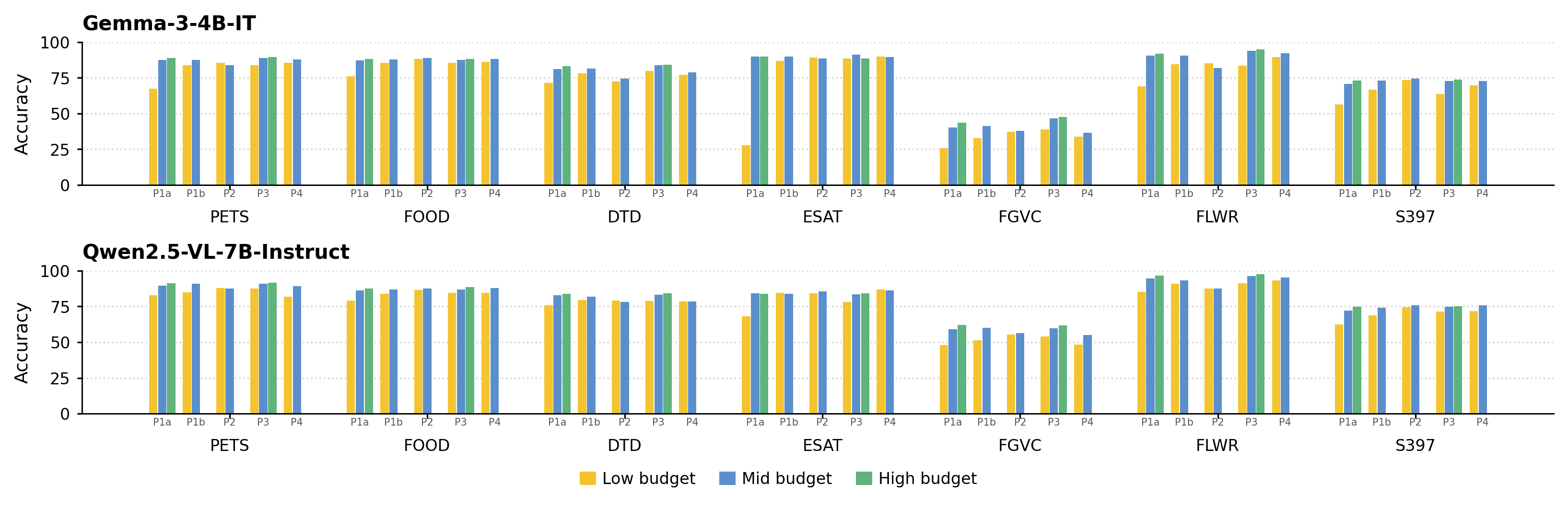}
  \vskip -3mm
  \caption{
    \textbf{Pipeline accuracy across token budgets and datasets.}
Mean per-class accuracy for the five inference pipelines (P1(a): full-resolution image ICL, P1(b): compressed-image ICL, P2: instruction ICL, P3: hybrid, P4: flip hybrid) on seven fine-grained benchmarks, evaluated under three per-query token budgets: Low ($\sim$1.5k), Mid ($\sim$3.3k), and High ($\sim$6.3k). P1(b) is reported at the Low and Mid budgets only. Top: Gemma-3-4B-IT. Bottom: Qwen2.5-VL-7B-Instruct. At matched token budgets, the hybrid P3 is competitive with, or stronger than, P1(a) on most datasets, while P2 at low budgets approaches mid-budget P3 on PETS, FOOD, ESAT, and FLWR. This is a per-dataset bar-chart view of the same results plotted as curves in Figure~\ref{fig:budget_line} in the main text.}
  \label{fig:budget_plot}
\end{figure*}


\begin{table*}[t]
\centering
\renewcommand{\arraystretch}{1.15}
\begin{tabular}{lccccccccc}
\toprule
\textbf{K} & \multicolumn{2}{c}{\textbf{Acc. P1}} & \textbf{Acc.} & \multicolumn{3}{c}{\textbf{Accuracy P3}} & \multicolumn{3}{c}{\textbf{Accuracy P4}} \\
\cmidrule(lr){2-3} \cmidrule(lr){5-7} \cmidrule(lr){8-10}
& P1(a) & P1(b) & P2 & $k'=1$ & $k'=2$ & $k'=3$ & $k'=1$ & $k'=2$ & $k'=3$ \\
\midrule
0  & 0.5286 & --     & --     & --     & --     & --     & --     & --     & --     \\
1  & 0.6728 & 0.663  & 0.7224 & 0.8016 & 0.8356 & 0.8388 & 0.8259 & 0.8508 & 0.8714 \\
2  & 0.7827 & 0.7725 & 0.8077 & 0.8555 & 0.8626 & 0.8713 & 0.8441 & 0.8619 & 0.8762 \\
3  & 0.8375 & 0.8366 & 0.8175 & 0.87 & 0.8697 & 0.8779 & 0.8541 & 0.8688 & 0.8772 \\
5  & 0.8742 & 0.8624 & 0.8276 & 0.8754 & 0.8858 & 0.8903 & 0.8521 & 0.8647 & 0.8778 \\
10 & 0.8888 & 0.8764 & 0.8545 & 0.8726     & 0.8962     & 0.8962     & 0.8629     & 0.8702     & 0.8799     \\
25 & --     & --     & 0.8390 & --     & --     & --     & --     & --     & --     \\
\bottomrule
\end{tabular}
\caption{Accuracy metrics for Gemma-3-4b-it across various shot counts ($K$) for PETS dataset. Acc. P1(a) is with full-resolution images in context, P1(b) uses compressed images.}
\label{tab:gemma-results}
\end{table*}

\begin{table*}[t]
\centering
\renewcommand{\arraystretch}{1.15}
\begin{tabular}{lccccccccc}
\toprule
\textbf{K} & \multicolumn{2}{c}{\textbf{Acc. P1}} & \textbf{Acc.} & \multicolumn{3}{c}{\textbf{Accuracy P3}} & \multicolumn{3}{c}{\textbf{Accuracy P4}} \\
\cmidrule(lr){2-3} \cmidrule(lr){5-7} \cmidrule(lr){8-10}
& P1(a) & P1(b) & P2 & $k'=1$ & $k'=2$ & $k'=3$ & $k'=1$ & $k'=2$ & $k'=3$ \\
\midrule
0  & 0.1103 & --     & --     & --     & --     & --     & --     & --     & --     \\
1  & 0.7627 & 0.752  & 0.7850 & 0.8154 & 0.8410 & 0.8537 & 0.8238 & 0.8464 & 0.8628 \\
2  & 0.8332 & 0.8292 & 0.8236 & 0.8520 & 0.8618 & 0.8693 & 0.8500 & 0.8626 & 0.8689 \\
3  & 0.8565 & 0.8536 & 0.8413 & 0.8650 & 0.8727 & 0.8788 & 0.8615 & 0.8687 & 0.8730 \\
5  & 0.8726 & 0.8694 & 0.8650 & 0.8671     & 0.8718     & 0.874     & 0.8712 & 0.8722 & 0.8760 \\
10 & 0.8818 & 0.8784 & 0.8814 & 0.8773     & 0.8813     & 0.8828     & 0.8794 & 0.8768 & 0.8817 \\
25 & --     & --     & 0.8903 & --     & --     & --     & --     & --     & --     \\
\bottomrule
\end{tabular}
\caption{Accuracy metrics for Gemma-3-4b-it across various shot counts ($K$) for FOOD dataset. Acc. P1(a) is with full-resolution images in context, P1(b) uses compressed images.}
\label{tab:gemma-results-food}
\end{table*}

\begin{table*}[t]
\centering
\renewcommand{\arraystretch}{1.15}
\begin{tabular}{lccccccccc}
\toprule
\textbf{K} & \multicolumn{2}{c}{\textbf{Acc. P1}} & \textbf{Acc.} & \multicolumn{3}{c}{\textbf{Accuracy P3}} & \multicolumn{3}{c}{\textbf{Accuracy P4}} \\
\cmidrule(lr){2-3} \cmidrule(lr){5-7} \cmidrule(lr){8-10}
& P1(a) & P1(b) & P2 & $k'=1$ & $k'=2$ & $k'=3$ & $k'=1$ & $k'=2$ & $k'=3$ \\
\midrule
0  & 0.1335 & --     & --     & --     & --     & --     & --     & --     & --     \\
1  & 0.7159 & 0.7028 & 0.7664 & 0.7452 & 0.7893 & 0.7968 & 0.7765 & 0.8015 & 0.7994 \\
2  & 0.7755 & 0.7601 & 0.7686 & 0.8175 & 0.8271 & 0.8313 & 0.7803 & 0.7920 & 0.7893 \\
3  & 0.7978 & 0.7815 & 0.7510 & 0.8180 & 0.8340 & 0.8335 & 0.7702 & 0.7787 & 0.7770 \\
5  & 0.8132 & 0.8027 & 0.7579 & 0.8218 & 0.8335 & 0.8404 & 0.7537 & 0.7824 & 0.7813 \\
10 & 0.8324 & 0.8145 & 0.7239 & 0.8138     & 0.834     & 0.8426     & 0.7597 & 0.7755 & 0.7888 \\
25 & --     & --     & 0.7425 & --     & --     & --     & --     & --     & --     \\
\bottomrule
\end{tabular}
\caption{Accuracy metrics for Gemma-3-4b-it across various shot counts ($K$) for DTD dataset. Acc. P1(a) is with full-resolution images in context, P1(b) uses compressed images.}
\label{tab:gemma-results-2}
\end{table*}

\begin{table*}[t]
\centering
\renewcommand{\arraystretch}{1.15}
\begin{tabular}{lccccccccc}
\toprule
\textbf{K} & \multicolumn{2}{c}{\textbf{Acc. P1}} & \textbf{Acc.} & \multicolumn{3}{c}{\textbf{Accuracy P3}} & \multicolumn{3}{c}{\textbf{Accuracy P4}} \\
\cmidrule(lr){2-3} \cmidrule(lr){5-7} \cmidrule(lr){8-10}
& P1(a) & P1(b) & P2 & $k'=1$ & $k'=2$ & $k'=3$ & $k'=1$ & $k'=2$ & $k'=3$ \\
\midrule
0  & 0.0437 & --     & --     & --     & --     & --     & --     & --     & --     \\
1  & 0.2781 & 0.2781 & 0.8571 & 0.8529 & 0.8759 & 0.8847 & 0.8811 & 0.8924 & 0.9000 \\
2  & 0.7900 & 0.7900 & 0.8715 & 0.8892 & 0.9066 & 0.9054 & 0.8906 & 0.9006 & 0.9049 \\
3  & 0.8690 & 0.8690 & 0.8892 & 0.8846 & 0.9074 & 0.9053 & 0.8995 & 0.9043 & 0.9094 \\
5  & 0.8996 & 0.8996 & 0.8894 & 0.8859 & 0.9035 & 0.9109 & 0.8901 & 0.8971 & 0.9078 \\
10 & 0.8991 & 0.8991 & 0.8921 & 0.8367 & 0.8738 & 0.8873 & 0.8790 & 0.8863 & 0.8952 \\
25 & --     & --     & 0.8868 & --     & --     & --     & --     & --     & --     \\
\bottomrule
\end{tabular}
\caption{Accuracy metrics for Gemma-3-4b-it across various shot counts ($K$) for ESAT dataset. ESAT images are already low-resolution, so Acc. P1(a) and P1(b) coincide.}
\label{tab:gemma-results-4}
\end{table*}

\begin{table*}[t]
\centering
\renewcommand{\arraystretch}{1.15}
\begin{tabular}{lccccccccc}
\toprule
\textbf{K} & \multicolumn{2}{c}{\textbf{Acc. P1}} & \textbf{Acc.} & \multicolumn{3}{c}{\textbf{Accuracy P3}} & \multicolumn{3}{c}{\textbf{Accuracy P4}} \\
\cmidrule(lr){2-3} \cmidrule(lr){5-7} \cmidrule(lr){8-10}
& P1(a) & P1(b) & P2 & $k'=1$ & $k'=2$ & $k'=3$ & $k'=1$ & $k'=2$ & $k'=3$ \\
\midrule
0  & 0.0792 & --     & --     & --     & --     & --     & --     & --     & --     \\
1  & 0.2565 & 0.2371 & 0.30153 & 0.34177 & 0.37143 & 0.3894 & 0.3288 & 0.3774 & 0.405  \\
2  & 0.31023 & 0.2885 & 0.33783 & 0.38793 & 0.4041  & 0.4119 & 0.3306 & 0.3717 & 0.3903 \\
3  & 0.36513 & 0.3277 & 0.34713 & 0.42184 & 0.43444 & 0.44344 & 0.3399 & 0.3706 & 0.3819 \\
5  & 0.40144 & 0.366  & 0.35823 & 0.447   & 0.4578  & 0.4647 & 0.3309 & 0.3675 & 0.3675 \\
10 & 0.43654 & 0.4122 & 0.37413 & 0.4776  & 0.4704  & 0.4755 & 0.3282 & 0.3405 & 0.3636 \\
25 & --     & --     & 0.37983 & --     & --     & --     & --     & --     & --     \\
\bottomrule
\end{tabular}
\caption{Accuracy metrics for Gemma-3-4b-it across various shot counts ($K$) for FGVC dataset. Acc. P1(a) is with full-resolution images in context, P1(b) uses compressed images.}
\label{tab:gemma-results-fgvc}
\end{table*}

\begin{table*}[t]
\centering
\renewcommand{\arraystretch}{1.15}
\begin{tabular}{lccccccccc}
\toprule
\textbf{K} & \multicolumn{2}{c}{\textbf{Acc. P1}} & \textbf{Acc.} & \multicolumn{3}{c}{\textbf{Accuracy P3}} & \multicolumn{3}{c}{\textbf{Accuracy P4}} \\
\cmidrule(lr){2-3} \cmidrule(lr){5-7} \cmidrule(lr){8-10}
& P1(a) & P1(b) & P2 & $k'=1$ & $k'=2$ & $k'=3$ & $k'=1$ & $k'=2$ & $k'=3$ \\
\midrule
0  & 0.1068 & --     & --     & --     & --     & --     & --     & --     & --     \\
1  & 0.6901 & 0.688  & 0.7921 & 0.7803 & 0.8186 & 0.8343 & 0.8882 & 0.9176 & 0.9264 \\
2  & 0.8284 & 0.813  & 0.8843 & 0.9    & 0.9058 & 0.9137 & 0.8931 & 0.9088 & 0.9225 \\
3  & 0.8715 & 0.845  & 0.8892 & 0.9225 & 0.9382 & 0.9421 & 0.8941 & 0.9078 & 0.9215 \\
5  & 0.9049 & 0.901  & 0.8715 & 0.9421 & 0.933  & 0.9382 & 0.8843 & 0.9186 & 0.9205 \\
10 & 0.9196 & 0.9057 & 0.8509 & 0.9421 & 0.945  & 0.949  & 0.892  & 0.9    & 0.9205 \\
25 & --     & --     & 0.8196 & --     & --     & --     & --     & --     & --     \\
\bottomrule
\end{tabular}
\caption{Accuracy metrics for Gemma-3-4b-it across various shot counts ($K$) for FLWR dataset. Acc. P1(a) is with full-resolution images in context, P1(b) uses compressed images.}
\label{tab:gemma-results-flwr}
\end{table*}

\begin{table*}[t]
\centering
\renewcommand{\arraystretch}{1.15}
\begin{tabular}{lccccccccc}
\toprule
\textbf{K} & \multicolumn{2}{c}{\textbf{Acc. P1}} & \textbf{Acc.} & \multicolumn{3}{c}{\textbf{Accuracy P3}} & \multicolumn{3}{c}{\textbf{Accuracy P4}} \\
\cmidrule(lr){2-3} \cmidrule(lr){5-7} \cmidrule(lr){8-10}
& P1(a) & P1(b) & P2 & $k'=1$ & $k'=2$ & $k'=3$ & $k'=1$ & $k'=2$ & $k'=3$ \\
\midrule
0  & 0.1925 & --     & --     & --     & --     & --     & --     & --     & --     \\
1  & 0.5618 & 0.5549 & 0.5817 & 0.5601 & 0.6016 & 0.639  & 0.6453 & 0.6837 & 0.6984 \\
2  & 0.6319 & 0.6406 & 0.6512 & 0.6446 & 0.6671 & 0.6925 & 0.6849 & 0.6926 & 0.7029 \\
3  & 0.67   & 0.6686 & 0.6756 & 0.6848 & 0.699  & 0.7087 & 0.6967 & 0.7053 & 0.7096 \\
5  & 0.7065 & 0.7092 & 0.7111 & 0.7136 & 0.7214 & 0.7266 & 0.7051 & 0.7127 & 0.7215 \\
10 & 0.73   & 0.7303 & 0.7338 & 0.7308 & 0.7367 & 0.7392 & 0.7203 & 0.7238 & 0.7282 \\
25 & --     & --     & 0.744  & --     & --     & --     & --     & --     & --     \\
\bottomrule
\end{tabular}
\caption{Accuracy metrics for Gemma-3-4b-it across various shot counts ($K$) for S397 dataset. Acc. P1(a) is with full-resolution images in context, P1(b) uses compressed images.}
\label{tab:gemma-results-s397}
\end{table*}


\begin{table*}[t]
\centering
\renewcommand{\arraystretch}{1.15}
\begin{tabular}{lccccccccc}
\toprule
\textbf{K} & \multicolumn{2}{c}{\textbf{Acc. P1}} & \textbf{Acc.} & \multicolumn{3}{c}{\textbf{Accuracy P3}} & \multicolumn{3}{c}{\textbf{Accuracy P4}} \\
\cmidrule(lr){2-3} \cmidrule(lr){5-7} \cmidrule(lr){8-10}
& P1(a) & P1(b) & P2 & $k'=1$ & $k'=2$ & $k'=3$ & $k'=1$ & $k'=2$ & $k'=3$ \\
\midrule
0  & 0.3571 & --     & --     & --     & --     & --     & --     & --     & --     \\
1  & 0.7915 & 0.7856 & 0.7969 & 0.8128 & 0.8336 & 0.8435 & 0.8253 & 0.8461 & 0.8601 \\
2  & 0.8329 & 0.8291 & 0.8368 & 0.8387 & 0.8422 & 0.8501 & 0.8383 & 0.8596 & 0.8671 \\
3  & 0.8466 & 0.8389 & 0.8472 & 0.8551 & 0.8563 & 0.8586 & 0.8467 & 0.8659 & 0.872  \\
5  & 0.8615 & 0.8454 & 0.8599 & 0.8687 & 0.8702 & 0.8703 & 0.8563 & 0.8719 & 0.875  \\
10 & 0.8759 & 0.8685 & 0.8669 & 0.8832 & 0.8837 & 0.8867 & 0.8578 & 0.8772 & 0.8795 \\
25 & --     & --     & 0.8744 & --     & --     & --     & --     & --     & --     \\
\bottomrule
\end{tabular}
\caption{Accuracy metrics for Qwen2.5-VL-7B-Instruct across various shot counts ($K$) for FOOD dataset. Acc. P1(a) is with full-resolution images in context, P1(b) uses compressed images.}
\label{tab:qwen-results}
\end{table*}

\begin{table*}[t]
\centering
\renewcommand{\arraystretch}{1.15}
\begin{tabular}{lccccccccc}
\toprule
\textbf{K} & \multicolumn{2}{c}{\textbf{Acc. P1}} & \textbf{Acc.} & \multicolumn{3}{c}{\textbf{Accuracy P3}} & \multicolumn{3}{c}{\textbf{Accuracy P4}} \\
\cmidrule(lr){2-3} \cmidrule(lr){5-7} \cmidrule(lr){8-10}
& P1(a) & P1(b) & P2 & $k'=1$ & $k'=2$ & $k'=3$ & $k'=1$ & $k'=2$ & $k'=3$ \\
\midrule
0  & 0.1672 & --     & --     & --     & --     & --     & --     & --     & --     \\
1  & 0.759  & 0.7218 & 0.7452 & 0.7648 & 0.776  & 0.7898 & 0.7712 & 0.7787 & 0.7905 \\
2  & 0.8    & 0.773  & 0.7735 & 0.7973 & 0.8042 & 0.8132 & 0.7877 & 0.7973 & 0.8026 \\
3  & 0.8085 & 0.7936 & 0.7835 & 0.8106 & 0.8212 & 0.8281 & 0.7856 & 0.8015 & 0.8048 \\
5  & 0.8266 & 0.7989 & 0.7957 & 0.8138 & 0.8239 & 0.8329 & 0.7745 & 0.7835 & 0.7888 \\
10 & 0.8378 & 0.8165 & 0.792  & 0.8308 & 0.8436 & 0.842  & 0.7643 & 0.7792 & 0.7835 \\
25 & --     & --     & 0.7824 & --     & --     & --     & --     & --     & --     \\
\bottomrule
\end{tabular}
\caption{Accuracy metrics for Qwen2.5-VL-7B-Instruct across various shot counts ($K$) for DTD dataset. Acc. P1(a) is with full-resolution images in context, P1(b) uses compressed images.}
\label{tab:qwen-results-dtd}
\end{table*}

\begin{table*}[t]
\centering
\renewcommand{\arraystretch}{1.15}
\begin{tabular}{lccccccccc}
\toprule
\textbf{K} & \multicolumn{2}{c}{\textbf{Acc. P1}} & \textbf{Acc.} & \multicolumn{3}{c}{\textbf{Accuracy P3}} & \multicolumn{3}{c}{\textbf{Accuracy P4}} \\
\cmidrule(lr){2-3} \cmidrule(lr){5-7} \cmidrule(lr){8-10}
& P1(a) & P1(b) & P2 & $k'=1$ & $k'=2$ & $k'=3$ & $k'=1$ & $k'=2$ & $k'=3$ \\
\midrule
0  & 0.6903 & --     & --     & --     & --     & --     & --     & --     & --     \\
1  & 0.8285 & 0.8077 & 0.8111 & 0.8574 & 0.8705 & 0.8743 & 0.8078 & 0.8323 & 0.846  \\
2  & 0.8552 & 0.8331 & 0.843  & 0.8757 & 0.8841 & 0.8852 & 0.8192 & 0.8601 & 0.8746 \\
3  & 0.8721 & 0.8504 & 0.8577 & 0.8887 & 0.892  & 0.8912 & 0.819  & 0.8708 & 0.886  \\
5  & 0.8972 & 0.8864 & 0.8778 & 0.9013 & 0.9117 & 0.9084 & 0.8252 & 0.8781 & 0.8917 \\
10 & 0.9132 & 0.9076 & 0.8803 & 0.9076 & 0.9198 & 0.9146 & 0.8323     & 0.8825     & 0.8939     \\
25 & --     & --     & 0.8768 & --     & --     & --     & --     & --     & --     \\
\bottomrule
\end{tabular}
\caption{Accuracy metrics for Qwen2.5-VL-7B-Instruct across various shot counts ($K$) for PETS dataset. Acc. P1(a) is with full-resolution images in context, P1(b) uses compressed images.}
\label{tab:qwen-results-pets}
\end{table*}

\begin{table*}[t]
\centering
\renewcommand{\arraystretch}{1.15}
\begin{tabular}{lccccccccc}
\toprule
\textbf{K} & \multicolumn{2}{c}{\textbf{Acc. P1}} & \textbf{Acc.} & \multicolumn{3}{c}{\textbf{Accuracy P3}} & \multicolumn{3}{c}{\textbf{Accuracy P4}} \\
\cmidrule(lr){2-3} \cmidrule(lr){5-7} \cmidrule(lr){8-10}
& P1(a) & P1(b) & P2 & $k'=1$ & $k'=2$ & $k'=3$ & $k'=1$ & $k'=2$ & $k'=3$ \\
\midrule
0  & 0.2413 & --     & --     & --     & --     & --     & --     & --     & --     \\
1  & 0.8529 & 0.84   & 0.8441 & 0.9147 & 0.9098 & 0.9127 & 0.9245 & 0.9421 & 0.946  \\
2  & 0.9107 & 0.899  & 0.8588 & 0.949  & 0.9509 & 0.949  & 0.9303 & 0.9441 & 0.951  \\
3  & 0.9264 & 0.9086 & 0.8725 & 0.9578 & 0.9617 & 0.9568 & 0.9334 & 0.948  & 0.9509 \\
5  & 0.947  & 0.9194 & 0.8784 & 0.9627 & 0.9637 & 0.9637 & 0.9313 & 0.945  & 0.9529 \\
10 & 0.9656 & 0.932  & 0.8745 & 0.9764 & 0.9745 & 0.9764 & 0.9225 & 0.9392 & 0.9529 \\
25 & --     & --     & 0.8745 & --     & --     & --     & --     & --     & --     \\
\bottomrule
\end{tabular}
\caption{Accuracy metrics for Qwen2.5-VL-7B-Instruct across various shot counts ($K$) for FLWR dataset. Acc. P1(a) is with full-resolution images in context, P1(b) uses compressed images.}
\label{tab:qwen-results-flwr}
\end{table*}

\begin{table*}[t]
\centering
\renewcommand{\arraystretch}{1.15}
\begin{tabular}{lccccccccc}
\toprule
\textbf{K} & \multicolumn{2}{c}{\textbf{Acc. P1}} & \textbf{Acc.} & \multicolumn{3}{c}{\textbf{Accuracy P3}} & \multicolumn{3}{c}{\textbf{Accuracy P4}} \\
\cmidrule(lr){2-3} \cmidrule(lr){5-7} \cmidrule(lr){8-10}
& P1(a) & P1(b) & P2 & $k'=1$ & $k'=2$ & $k'=3$ & $k'=1$ & $k'=2$ & $k'=3$ \\
\midrule
0  & 0.2826 & --     & --     & --     & --     & --     & --     & --     & --     \\
1  & 0.4794 & 0.4449 & 0.4392 & 0.4967 & 0.5265 & 0.5391 & 0.4689 & 0.5163 & 0.5433 \\
2  & 0.5345 & 0.4875 & 0.4569 & 0.5313 & 0.5616 & 0.5535 & 0.4752 & 0.5304 & 0.5409 \\
3  & 0.5598 & 0.5139 & 0.4784 & 0.5709 & 0.5778 & 0.5766 & 0.4836 & 0.5286 & 0.543  \\
5  & 0.5916 & 0.5584 & 0.5148 & 0.5892 & 0.5979 & 0.5955 & 0.4926 & 0.5397 & 0.5442 \\
10 & 0.6222 & 0.5988 & 0.5537 & 0.5973 & 0.6192 & 0.6174 & 0.5215 & 0.5457 & 0.549  \\
25 & --     & --     & 0.5627 & --     & --     & --     & --     & --     & --     \\
\bottomrule
\end{tabular}
\caption{Accuracy metrics for Qwen2.5-VL-7B-Instruct across various shot counts ($K$) for FGVC dataset. Acc. P1(a) is with full-resolution images in context, P1(b) uses compressed images.}
\label{tab:qwen-results-fgvc}
\end{table*}

\begin{table*}[t]
\centering
\renewcommand{\arraystretch}{1.15}
\begin{tabular}{lccccccccc}
\toprule
\textbf{K} & \multicolumn{2}{c}{\textbf{Acc. P1}} & \textbf{Acc.} & \multicolumn{3}{c}{\textbf{Accuracy P3}} & \multicolumn{3}{c}{\textbf{Accuracy P4}} \\
\cmidrule(lr){2-3} \cmidrule(lr){5-7} \cmidrule(lr){8-10}
& P1(a) & P1(b) & P2 & $k'=1$ & $k'=2$ & $k'=3$ & $k'=1$ & $k'=2$ & $k'=3$ \\
\midrule
0  & 0.1164 & --     & --     & --     & --     & --     & --     & --     & --     \\
1  & 0.6797 & 0.6797 & 0.8521 & 0.752  & 0.7875 & 0.7816 & 0.8759 & 0.8841 & 0.883  \\
2  & 0.8217 & 0.8217 & 0.8486 & 0.8218 & 0.8156 & 0.8141 & 0.8789 & 0.8878 & 0.8885 \\
3  & 0.8452 & 0.8452 & 0.8529 & 0.8402 & 0.828  & 0.8277 & 0.87   & 0.8941 & 0.8915 \\
5  & 0.8406 & 0.8406 & 0.8519 & 0.8362 & 0.844  & 0.8351 & 0.8563 & 0.8778 & 0.8785 \\
10 & 0.8383 & 0.8383 & 0.8427 & 0.8537 & 0.8481 & 0.8422 & 0.8293 & 0.8615 & 0.863  \\
25 & --     & --     & 0.8541 & --     & --     & --     & --     & --     & --     \\
\bottomrule
\end{tabular}
\caption{Accuracy metrics for Qwen2.5-VL-7B-Instruct across various shot counts ($K$) for ESAT dataset. ESAT images are already low-resolution, so Acc. P1(a) and P1(b) coincide.}
\label{tab:qwen-results-esat}
\end{table*}

\begin{table*}[t]
\centering
\renewcommand{\arraystretch}{1.15}
\begin{tabular}{lccccccccc}
\toprule
\textbf{K} & \multicolumn{2}{c}{\textbf{Acc. P1}} & \textbf{Acc.} & \multicolumn{3}{c}{\textbf{Accuracy P3}} & \multicolumn{3}{c}{\textbf{Accuracy P4}} \\
\cmidrule(lr){2-3} \cmidrule(lr){5-7} \cmidrule(lr){8-10}
& P1(a) & P1(b) & P2 & $k'=1$ & $k'=2$ & $k'=3$ & $k'=1$ & $k'=2$ & $k'=3$ \\
\midrule
0  & 0.2137 & --     & --     & --     & --     & --     & --     & --     & --     \\
1  & 0.6255 & 0.6163 & 0.6233 & 0.6803 & 0.6996 & 0.7157 & 0.6685 & 0.6868 & 0.6977 \\
2  & 0.6755 & 0.6544 & 0.6813 & 0.7022 & 0.7114 & 0.7228 & 0.7009 & 0.7151 & 0.7238 \\
3  & 0.6986 & 0.6865 & 0.7048 & 0.7179 & 0.7269 & 0.7324 & 0.7186 & 0.73   & 0.7361 \\
5  & 0.7202 & 0.7089 & 0.7247 & 0.7256 & 0.7348 & 0.7475 & 0.7376 & 0.7465 & 0.7516 \\
10 & 0.7483 & 0.741  & 0.7438 & 0.7398 & 0.7457 & 0.7521 & 0.7486 & 0.7534 & 0.7586 \\
25 & --     & --     & 0.7572 & --     & --     & --     & --     & --     & --     \\
\bottomrule
\end{tabular}
\caption{Accuracy metrics for Qwen2.5-VL-7B-Instruct across various shot counts ($K$) for S397 dataset. Acc. P1(a) is with full-resolution images in context, P1(b) uses compressed images.}
\label{tab:qwen-results-s397}
\end{table*}

\clearpage
\twocolumn

\section{Full Prompt Templates}
\label{sec:appendix_prompts}

\subsection*{E.1\quad Instruction Generation (offline prerequisite)}

\begin{tcolorbox}[promptbox,
  colback=promptbg-instgen, colframe=promptframe-instgen,
  title={\small\bfseries\color{white} Instruction Generation Prompt (system + user, \textit{image attached})}]
  You are an expert in food recognition and culinary visual analysis. Your task is to generate generalizable identification rules for a food category from the Food-101 dataset that can be applied to ANY image of this dish — not just the example shown.

    Food Category: \{label\}\\

    Generate exactly 3 identification rules in the following format:
    \begin{enumerate}
    \item  To identify [food category], look for [general visual cue — shape, color, dominant ingredients, or plating style]
    \item  To distinguish from similar dishes, focus on [differentiating characteristic — surface texture, garnish, sauce, layering, or cooking method cues such as browning, glaze, or char]
    \item  To avoid misclassification, check [common confusion point with a visually similar dish and how to resolve it]
    \end{enumerate}

    Rules must be:
    \begin{itemize}
    \item  General enough to apply to any image of {label}, not just this one
    \item  Focused on visually observable properties (color, shape, surface texture, ingredient arrangement, portion style, cooking cues)
    \item  Concise one sentence each
    \end{itemize}

    Do not write paragraphs. Do not describe this specific image. Do not ask questions.
\end{tcolorbox}

\subsection*{E.2\quad Zero-Shot Inference (P0)}

\begin{tcolorbox}[promptbox,
  colback=promptbg-zero, colframe=promptframe-zero,
  title={\small\bfseries\color{white} Zero-Shot Prompt (user turn, \textit{image attached})}]
    Identify the food dish shown in the image.
    Provide only the food category name from the Food-101 dataset. Do not output anything else.

\end{tcolorbox}

\subsection*{E.3\quad Image ICL (Same for P1(a) and P1(b))}

\begin{tcolorbox}[promptbox,
  colback=promptbg-image, colframe=promptframe-image,
  title={\small\bfseries\color{white} P1: System Message}]
    You are an expert food recognition system trained on visual properties of diverse cuisines and dishes. Study the following example images carefully.
\end{tcolorbox}

\begin{tcolorbox}[promptbox,
  colback=promptbg-image, colframe=promptframe-image,
  title={\small\bfseries\color{white} P1: Per-Example Block (repeated $K$ times, \textit{image attached})}]
    Example \{idx\}:\\
    Category: \{label\}
\end{tcolorbox}

\begin{tcolorbox}[promptbox,
  colback=promptbg-image, colframe=promptframe-image,
  title={\small\bfseries\color{white} P1: Target Prompt (\textit{query image attached})}]
    Now classify the food dish in the following image.
    Provide only the food category name. Do not output anything else.
\end{tcolorbox}

\subsection*{E.4\quad Instruction ICL (P2)}

\begin{tcolorbox}[promptbox,
  colback=promptbg-instr, colframe=promptframe-instr,
  title={\small\bfseries\color{white} P2: System Message}]
    You are an expert food classifier. Below are reasoning instructions derived from Food-101 images that are visually similar to the target image. Go through them carefully in order to classify the target image.

\end{tcolorbox}

\begin{tcolorbox}[promptbox,
  colback=promptbg-instr, colframe=promptframe-instr,
  title={\small\bfseries\color{white} P2: Per-Instruction Block (repeated $K$ times, text only)}]
    Instruction \{idx\}:\\
    Category: \{label\}\\
    Instruction: \{instruction\}
\end{tcolorbox}

\begin{tcolorbox}[promptbox,
  colback=promptbg-instr, colframe=promptframe-instr,
  title={\small\bfseries\color{white} P2: Target Prompt (\textit{query image attached})}]
    Apply the above reasoning strategies to classify the following image into a Food-101 category name. Focus on visible dish properties like ingredient composition, color, shape, cooking method cues, and plating style and relate them to the instructions provided.
    Provide only the food category name, similar to what is mentioned in the instructions provided. 

Question: \{question\}\\
Answer:
\end{tcolorbox}

\subsection*{E.5\quad Hybrid ICL: Image-Primary (P3)}

\begin{tcolorbox}[promptbox,
  colback=promptbg-hybrid, colframe=promptframe-hybrid,
  title={\small\bfseries\color{white} P3: System Message}]
 You are an expert in fine-grained food classification. You will be given two types of in-context examples:
 \begin{enumerate}
    \item  Visual examples: images paired with their correct food category name.
    \item  Reasoning instructions: text-based strategies derived from visually similar Food-101 images, describing how to identify the food category.
    \end{enumerate}
    Your goal is to correctly identify the food category of the target image. Study all examples and instructions carefully before answering.

\end{tcolorbox}

\vspace{-5mm}
\begin{tcolorbox}[promptbox,
  colback=promptbg-hybrid, colframe=promptframe-hybrid,
  title={\small\bfseries\color{white} P3: Visual Example Block (repeated $K$ times, \textit{image attached})}]
    Example \{idx\}:\\
    Category: \{label\}
\end{tcolorbox}

\vspace{-5mm}
\begin{tcolorbox}[promptbox,
  colback=promptbg-hybrid, colframe=promptframe-hybrid,
  title={\small\bfseries\color{white} P3: Reasoning Instruction Block (repeated $K'$ times, text only)}]
    Instruction Example \{idx\}:\\
    Category: \{label\}\\
    Instruction: \{instruction\}
\end{tcolorbox}

\vspace{-2pt}
\begin{tcolorbox}[promptbox,
  colback=promptbg-hybrid, colframe=promptframe-hybrid,
  title={\small\bfseries\color{white} P3: Target Prompt (\textit{query image attached})}]
    Now predict the food category name using both the visual examples and reasoning strategies above. Provide only the food category name, do not output anything else. Use the visual examples and the instructions to answer based on the target image only. Do not get confused with the example images.
    Give the exact food category name. Do not give a generic description. 

\end{tcolorbox}

\subsection*{E.6\quad Flip Hybrid ICL: Instruction-Primary (P4)}

\begin{tcolorbox}[promptbox,
  colback=promptbg-flip, colframe=promptframe-flip,
  title={\small\bfseries\color{white} P4: System Message}]
    You are an expert in fine-grained food classification. You will be given two types of in-context examples:
    \begin{enumerate}
    \item  Reasoning instructions: text-based strategies derived from visually similar images from the Food-101 dataset, describing how to identify the food category.
    \item  Visual examples: images paired with their correct food category name.
    \end{enumerate}
    Your primary guidance comes from the reasoning instructions. Use the visual examples as supplementary evidence. Study all instructions and examples carefully before answering.

\end{tcolorbox}

\begin{tcolorbox}[promptbox,
  colback=promptbg-flip, colframe=promptframe-flip,
  title={\small\bfseries\color{white} P4: Reasoning Instruction Block (repeated $K$ times, text only)}]
    Instruction Example \{idx\}:\\
    Category: \{label\}\\
    Instruction: \{instruction\}
\end{tcolorbox}

\begin{tcolorbox}[promptbox,
  colback=promptbg-flip, colframe=promptframe-flip,
  title={\small\bfseries\color{white} P4: Visual Example Block (repeated $K'$ times, \textit{image attached})}]
    Example \{idx\}:\\
    Category: \{label\}
\end{tcolorbox}

\vspace{4pt}
\begin{tcolorbox}[promptbox,
  colback=promptbg-flip, colframe=promptframe-flip,
  title={\small\bfseries\color{white} P4: Target Prompt (\textit{query image attached})}]
    Now predict the food category name using the reasoning instructions and visual examples above. Provide only the food category name, do not output anything else. Apply the reasoning strategies first, then use the visual examples to confirm. Do not get confused with the example images.
    Give the exact food category name. Do not give a generic description. 

\end{tcolorbox}

\section{Instruction Quality}
\label{sec:appendix_quality}
We used Gemma4-31b-it an LLM-judge to quantitatively evaluate instruction quality across 100 random images per dataset on three criteria: groundedness (1-5, higher means features mentioned are visible in the specific image), genericness (1-5, higher means the instruction is tailored to that specific image rather than reusable across any image of the class), and hallucination (1-5, higher means fewer fabricated features). Across all seven datasets, mean groundedness is 4.57 and mean hallucination is 3.87, indicating that the large majority of generated instructions reference real, visible features with relatively few fabricated details. Genericness is comparatively lower at a mean of 3.2, suggesting instructions lean moderately toward reusable, class-level phrasing rather than being tightly specific to the individual image, consistent with the label-prior tendency we identify and quantify separately via the label-only control experiment below.

\begin{table}[t]
  \centering
  \small
    \begin{tabular}{lccc}
        \toprule
        Dataset & Groundedness & Genericness & Hallucination \\
        \midrule
        PETS & 4.67 & 3.03 & 3.92 \\
        FOOD & 4.82 & 3.01 & 3.47 \\
        DTD  & 4.84 & 3.22 & 4.66 \\
        FGVC & 4.78 & 3.54 & 4.42 \\
        FLWR & 4.80 & 3.26 & 4.53 \\
        S397 & 4.81 & 3.00 & 3.48 \\
        ESAT         & 3.30 & 3.34 & 2.63 \\
        \midrule
        \textbf{Mean }        & \textbf{4.57} & \textbf{3.20} & \textbf{3.87} \\
        \bottomrule
    \end{tabular}
  \caption{\textbf{Instruction Quality.} Mean scores given by LLM as a judge for various datasets acrross different parameters. }
  \label{tab:instruction_quality}
\end{table}
Performance is not uniform across datasets. FGVC, DTD, FLWR, and FOOD score highest on groundedness (4.78-4.84); DTD, FGVC, and FLWR also score highest on hallucination (4.42-4.66), indicating the generator MLLM reliably describes real, visible features for object-centric categories with clear parts and textures. ESAT is a clear outlier, with the lowest groundedness (3.30) and lowest hallucination score (2.63), meaning satellite land-use images produce comparatively more ungrounded or fabricated descriptions. We attribute this to the abstract, low-texture nature of satellite imagery, where visual cues (spectral tone, coarse texture) are harder to verbalize precisely than object-centric features, making the generator more prone to defaulting to generic or invented class-level descriptions. The images have little contrasting features and hence the instructions have a low generic score. On manual evaluation, the MLLM labels green patches as forest land or grasslands, based on the class label, which is why we see a low hallucination score.
\begin{tcolorbox}[promptbox,
  colback=promptbg-instgen, colframe=promptframe-instgen,
  title={\small\bfseries\color{white} LLM Judge prompt to evaluate quality of instructions}]
  Food class: \{class name\}

Identification instruction:
\{instruction\}

Please evaluate this instruction on the following three criteria. Score each on a 1–5 integer scale.
\begin{enumerate}

\item  Groundedness (1–5)
   Does the instruction reference features that are actually visible in THIS specific image?
   1 = Completely ungrounded (references features invisible or absent in this image)
   5 = Perfectly grounded (every feature mentioned is clearly visible in this image)

\item  Genericness (1–5)
   Could this instruction apply equally well to ANY image of \{class name\}, or is it tailored to THIS image?
   1 = Completely generic (could be copy-pasted for any image of this class)
   5 = Highly specific (references details unique to this particular image)

\item  Hallucination (1–5)
   Does the instruction mention features NOT present in the image?
   1 = Severe hallucination (multiple features mentioned that are clearly absent)
   5 = No hallucination (every mentioned feature exists in the image)
\end{enumerate}
Respond ONLY with a valid JSON object in this exact format (no markdown, no extra text):
\{\{"groundedness": <int>, "genericness": <int>, "hallucination": <int>, "reasoning": "<one sentence explaining the scores>"\}\}
\end{tcolorbox}
We manually validated 50 of the LLM-judge scores per dataset against human ratings on the same three criteria, in order to confirm the automatic scores are not simply noise. During this validation, we observed a consistent pattern: instructions receiving low hallucination scores (i.e., more fabricated content) also tended to receive low genericness scores (i.e., more template-like, reusable phrasing), rather than being appearing as specific but wrong. This suggests that hallucination in our pipeline is primarily a failure of grounding (falling back to label-prior content) rather than a failure of invention (fabricating plausible-sounding but false specifics), which is a meaningfully less severe failure mode for a classification-support use case, since generic-but-wrong instructions are more likely to be simply uninformative for a given query rather than actively misleading.

\section{Ablation Study}
\label{sec:appendix_ablation}
To validate the key design choices in our instruction-based in-context learning pipeline, we conduct two targeted ablation studies: one on the \textbf{format} of the generated instructions, and one on the \textbf{granularity} at which they are produced. All ablations are evaluated on the Oxford-IIIT Pet and FGVC Aircraft dataset  test split using CLIP-based retrieval with Gemma model. We report Top-1 accuracy averaged across $K$ values. Results are available in Table \ref{tab:appendix_ablation}.

\subsection{Ablation A: Instruction Format}
Our baseline uses a structured 3-part instruction format, explicitly designed to cover three complementary aspects of visual identification: (1) a general appearance cue, (2) a discriminative feature separating the breed from visually similar ones, and (3) a common misclassification point with a resolution strategy. We ablate whether this structured format is necessary, or whether simpler textual descriptions of the same breed yield comparable guidance.
We compare four instruction formats:
\begin{itemize}[noitemsep]
\item Structured (Ours): Three-part rule following the fixed format described above.
\item Simple Caption: This tests whether a minimal, factual description suffices.
\item Unstructured Paragraph: A free-form 2–3 sentence description covering coat, face, body, and distinctive markings, with no imposed structure. This tests whether richer natural language compensates for the lack of explicit discriminative framing.
\item Single Rule: Only the discriminative rule, omitting the general and error-avoidance components. This isolates the contribution of inter-class discrimination alone.
\item Two Rule: Only the general appearance and discriminative rule, omitting the common confusion point. This isolates the contribution of the confusion point and highlights its necessity.
\end{itemize}
\textbf{The structured 3-part format outperforms all alternatives across all values of $K$}. The simple caption degrades performance, as a one-sentence description lacks sufficient specificity to distinguish between breeds that share broad visual characteristics (e.g., tabby cats or golden-coated dogs). The single-rule variant improves over the caption, suggesting that a formal instruction with inter-class discriminiation allows the model to distinguish much easily. The unstructured paragraph improves over the single-rule but still falls short of the baseline, suggesting that free-form descriptions, while richer, do not consistently surface the discriminative cues, and need a systematic representation and the general identification cue and the misclassification-avoidance rule provide complementary signal that the single rule cannot replicate alone.

\begin{tcolorbox}[promptbox,
  colback=promptbg-hybrid, colframe=promptframe-hybrid,
  title={\small\bfseries\color{white} Simple Caption Generation Prompt}]
      You are an animal identification assistant. Generate a simple one-sentence caption for the pet breed shown in the image.

    Pet Breed: \{label\}

    Write a single sentence in the format: "This is a [breed], which typically has [2-3 key visual features]."

    Do not write more than one sentence.
\end{tcolorbox}
\vspace{-4mm}
\begin{tcolorbox}[promptbox,
  colback=promptbg-hybrid, colframe=promptframe-hybrid,
  title={\small\bfseries\color{white} Single rule Generation Prompt}]
    You are an expert in animal morphology and veterinary visual identification. Generate a single discriminative identification rule for the pet breed shown in the image.

    Pet Breed: \{label\}

    Generate exactly 1 rule in the following format:
    To distinguish [breed] from similar breeds, focus on [key differentiating visual characteristic].

    The rule must be concise, one sentence. Focus on the most visually distinctive feature that separates this breed from others it is commonly confused with.

    Do not write more than one rule.
\end{tcolorbox}
\begin{tcolorbox}[promptbox,
  colback=promptbg-hybrid, colframe=promptframe-hybrid,
  title={\small\bfseries\color{white} Unstructured Paragraph Generation Prompt}]
   You are an expert in animal morphology. Describe the visual appearance of the pet breed shown in the image in a short paragraph.

    Pet Breed: \{label\}

    Write 2-3 natural sentences covering coat color and texture, face structure, body proportions, and any distinctive markings observable in pet photography. Be descriptive and specific to this breed.

    Do not use bullet points or numbered lists. 
\end{tcolorbox}

\subsection{Ablation (b): Instruction Granularity (Per-Image vs. Per-Class)}
In our pipeline, instructions are generated independently for each training image, meaning retrieved neighbors contribute image-specific guidance. An alternative is to generate a single canonical instruction per class, which would be shared across all retrieved images of that class. Per-class instructions are appealing because they could provide more stable, generalizable descriptions, and are cheaper to generate (one call per class rather than one per image).

For per-class instruction generation, we select a single representative image per class, the training image whose CLIP embedding is closest to the mean embedding of its class (the centroid representative). A structured 3-part instruction is generated for this representative image and stored as the canonical description for the entire class. At inference time, retrieval proceeds identically (top-K by CLIP cosine similarity), but after retrieval, examples are deduplicated by class and each unique class's canonical instruction is used in the context, rather than the per-image instruction of the retrieved neighbor.

After evaluation, we see that \textbf{per-class instructions underperform per-image instructions consistently across all $K$ values}. You can see the results in Table \ref{tab:appendix_ablation}. We attribute this to two compounding factors. First, representativeness loss: even the centroid image captures only one instantiation of a breed, and the instruction generated from it may not generalise to the full visual diversity of the class (different poses, lighting, angles, age). Second, retrieval–instruction mismatch: in the per-image setting, the retrieved neighbor and its instruction are aligned. The instruction was generated from a visually similar image, so its guidance is implicitly calibrated to the region of visual space where the test image lies. The per-class instruction breaks this alignment; the same instruction is applied regardless of which specific image was retrieved, losing the localised guidance that makes per-image retrieval effective. Together, these results suggest that the richness and diversity of per-image instructions is not redundant, it is a feature of the design that scales gracefully with the visual heterogeneity of the dataset.
\subsection{Ablation (c): Potential label-prior confound}
A possible confound in our design is that the generator MLLM may lean heavily on the class label itself when producing an instruction, rather than on visual evidence in the specific training image. If this were the dominant effect, per-image instructions would offer little over instructions conditioned on the label alone, and the gains we attribute to instruction distillation would instead be an artifact of the generator's label-conditioned priors.

To test this, we construct a \textbf{label-only control}: instructions generated with the identical prompt template (Appendix~\ref{sec:appendix_prompts}) and the identical three-part structure, but with the training image removed from the input, so the generator conditions only on the class label. We then run instruction ICL on FGVC and Oxford-IIIT Pets with per-image instructions and with label-only instructions under an otherwise matched pipeline (same retrieval index, same $K$ values, same backbone), and additionally compare the two instruction pools directly along several lexical and semantic diversity axes.

\textbf{Accuracy.} Across every tested $K \in \{1,2,3,5,10,25\}$ on both datasets, per-image instructions never underperform the label-only control; the ordering is consistent and directionally uniform rather than driven by a single favorable $K$ (Table~\ref{tab:label_prior_ablation}). On FGVC the gain is stable at 1.3--1.8pp across the full $K$ range, while on Oxford-IIIT Pets the gain grows from 0.5pp at $K{=}1$ to 3.7pp at $K{=}25$. This is the first piece of evidence against the label-prior explanation: if the generator were merely reciting label-conditioned boilerplate, we would expect the two conditions to perform comparably, or for the ordering to be inconsistent across $K$ and datasets.

\begin{table}[t]
  \centering
  \small
  \resizebox{\linewidth}{!}{
  \begin{tabular}{lccc}
    \toprule
    \textbf{K} & \textbf{Label-only} & \textbf{Per-image} & \textbf{$\Delta$ (pp)} \\
    \midrule
    1  & 0.7172 & 0.7224 & +0.5 \\
    2  & 0.7922 & 0.8077 & +1.6 \\
    3  & 0.7991 & 0.8175 & +1.8 \\
    5  & 0.8072 & 0.8276 & +2.0 \\
    10 & 0.8272 & 0.8545 & +2.7 \\
    25 & 0.8023 & 0.8390 & +3.7 \\
    \bottomrule
  \end{tabular}
  }
  \resizebox{\linewidth}{!}{
  \begin{tabular}{lccc}
    \toprule
    \textbf{K} & \textbf{Label-only} & \textbf{Per-image} & \textbf{$\Delta$ (pp)} \\
    \midrule
    1  & 0.2870 & 0.3015 & +1.5 \\
    2  & 0.3245 & 0.3378 & +1.3 \\
    3  & 0.3328 & 0.3471 & +1.4 \\
    5  & 0.3407 & 0.3582 & +1.8 \\
    10 & 0.3602 & 0.3741 & +1.4 \\
    25 & 0.3659 & 0.3798 & +1.4 \\
    \bottomrule
  \end{tabular}
  }
  \caption{\textbf{Label-prior control.} Top-1 accuracy for label-only vs.\ per-image instructions on Oxford-IIIT Pets (top) and FGVC (bottom) across $K \in \{1,2,3,5,10,25\}$. Per-image instructions outperform the label-only control at every $K$ on both datasets, with no exceptions.}
  \label{tab:label_prior_ablation}
\end{table}

\textbf{Diversity metrics.} The accuracy comparison alone cannot distinguish "the image adds useful signal" from "the image adds noise that happens not to hurt accuracy," so we directly measure whether the training image changes the \emph{content} of the generated instructions. We compute four metrics over instructions grouped by class:
\begin{itemize}[noitemsep]
  \item \textbf{Within-class cosine similarity}: mean pairwise cosine similarity between instruction embeddings of the same class. Lower values indicate the instructions for a class are less redundant with one another. Per-image instructions score lower (0.1424) than label-only instructions (0.1873).
  \item \textbf{Type-token ratio (TTR)}: the ratio of unique word types to total word tokens, a standard measure of vocabulary breadth. Per-image instructions have a higher TTR (0.0038 vs. 0.0029), indicating a broader working vocabulary across the instruction pool.
  \item \textbf{Unique trigram ratio}: the fraction of word trigrams that appear only once across the instruction pool. Per-image instructions score 0.0497 versus 0.0236 for label-only, roughly $2.1\times$ higher, indicating more varied phrase-level structure rather than reuse of the same stock phrases.
  \item \textbf{Exact duplicate ratio}: the fraction of instructions within a class that are byte-identical to another instruction in that class. This is the starkest signal: 0.017 for per-image instructions versus 0.653 for label-only instructions. With the image removed, the generator produces exactly the same instruction for a given class in roughly two-thirds of cases; grounding on a specific image reduces this collapse to under 2\%.
\end{itemize}

Taken together, these four metrics show that the training image is not a redundant input that the generator ignores in favor of the label: removing it causes the generator to collapse toward a small number of generic, class-templated outputs, while conditioning on the image consistently elicits distinct, image-specific phrasing. This is consistent with the LLM-judge \emph{genericness} scores reported in Appendix~\ref{sec:appendix_quality} (mean 3.2 on a 1--5 scale, lower than groundedness or hallucination scores), which independently suggested a moderate label-prior tendency in the generated instructions even in the per-image condition; the label-only control here isolates and quantifies that tendency directly, and confirms that per-image conditioning substantially, though not completely, counteracts it. Combined with the accuracy results, this rules out the possibility that the gains reported for instruction ICL in the main text are an artifact of label-conditioned descriptions or of sampling noise, and supports our design choice of generating instructions per training image rather than per class or per label.
\section{Efficiency Analysis}
\label{sec:appendix_efficiency}
We benchmarked wall-clock latency and throughput (100 queries per configuration, matched query samples, single A100 GPU) for all inference pipelines (Table~\ref{tab:pipeline_comparison}) to validate that the token-count reductions reported in Table~\ref{tab:token_cost} translate into real inference speedups. At K=5, the configuration underlying our reported 2.9x token reduction, mean latency drops from 0.5486s per query for image ICL (P1) to 0.1678s per query for instruction ICL (P2), a 3.27x speedup, exceeding the 2.9x token-count ratio. Throughput improves from 2.57 QPS (P1) to 24.40 QPS (P2), a 9.49x improvement. At K=1, latency drops from 0.2444s to 0.1405s (1.74x), and at K=10, from 1.1492s to 0.2112s (5.44x), showing the wall-clock advantage of instruction ICL grows with K, consistent with our finding that instruction ICL scales more favorably within a fixed token budget.

\begin{table}[t]
\centering
\small
\begin{tabular}{cccccc}
\toprule
\multirow{2}{*}{\textbf{K}} & \multirow{2}{*}{\textbf{Pipeline}} & \multirow{2}{*}{\textbf{Latency (s)}} & \multirow{2}{*}{\textbf{QPS}} & \multicolumn{2}{c}{\textbf{Speedup}} \\
\cmidrule(lr){5-6}
& & & & \textbf{Latency} & \textbf{QPS} \\
\midrule
1  & P1 & 0.2444 & 17.68 & ---   & ---   \\
1  & P2 & 0.1405 & 26.35 & 1.74x & 1.49x \\
5  & P1 & 0.5486 & 2.57  & ---   & ---   \\
5  & P2 & 0.1678 & 24.40 & 3.27x & 9.49x \\
10 & P1 & 1.1492 & 1.18  & ---   & ---   \\
10 & P2 & 0.2112 & 6.48  & 5.44x & 5.49x \\
\bottomrule
\end{tabular}
\caption{Latency and QPS comparison between P1 and P2 pipelines across different values of K.}
\label{tab:pipeline_comparison}
\end{table}

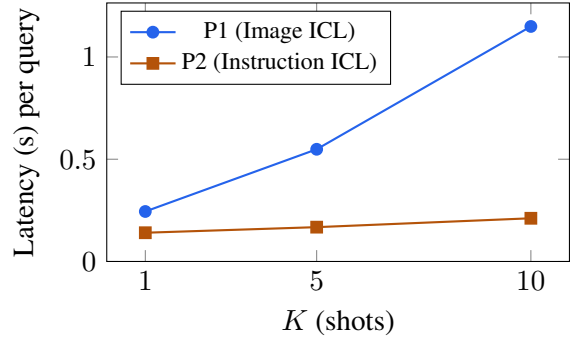
\begin{figure}[t]
\centering
\begin{tikzpicture}
\begin{axis}[
    width=\columnwidth,
    height=5cm,
    xlabel={$K$ (shots)},
    ylabel={Latency (s) per query},
    xtick={1,5,10},
    ymin=0,
    legend pos=north west,
    legend style={font=\small},
]
\addplot[color=promptframe-image, mark=*, thick] coordinates {(1,0.2444) (5,0.5486) (10,1.1492)};
\addplot[color=promptframe-instr, mark=square*, thick] coordinates {(1,0.1405) (5,0.1678) (10,0.2112)};
\legend{P1 (Image ICL), P2 (Instruction ICL)}
\end{axis}
\end{tikzpicture}
\caption{Per-query latency vs.\ shot count $K$ for P1 and P2 (Table~\ref{tab:pipeline_comparison}). P1's latency grows steeply with $K$ since each additional shot adds a full image ($\approx$550 tokens); P2's grows far more slowly, so the speedup widens from 1.74$\times$ at $K{=}1$ to 5.44$\times$ at $K{=}10$.}
\label{fig:latency_vs_k}
\end{figure}

A token-matched comparison isolates the modality effect from the token-count effect directly. At the Mid budget (P1 K=5: 3,400 tokens; P2 K=25: 3,305 tokens), P2's latency (0.3327s) is 39.4\% lower than P1's (0.5486s) despite near-identical token counts. This shows the speedup is not solely a consequence of using fewer tokens; it also reflects a difference in per-token processing cost between modalities. Total token-throughput (input+output TPS, Table~\ref{tab:token_budget_comparison}) supports this: at K=5, P2 processes tokens at 5,251.6 tokens/sec versus P1's 3,053.9 tokens/sec, a 1.72x higher raw rate. Since image tokens are processed more slowly than text tokens, the observed latency speedup (3.27x at K=5) exceeds what the token-count ratio (2.9x) alone would predict.

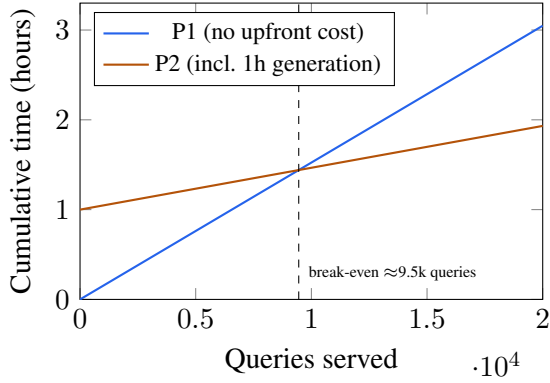
\begin{figure}[t]
\centering
\begin{tikzpicture}
\begin{axis}[
    width=\columnwidth,
    height=5.5cm,
    xlabel={Queries served},
    ylabel={Cumulative time (hours)},
    xmin=0, xmax=20000,
    ymin=0, ymax=3.3,
    legend pos=north west,
    legend style={font=\small},
]
\addplot[color=promptframe-image, thick, domain=0:20000, samples=100] {0.5486*x/3600};
\addplot[color=promptframe-instr, thick, domain=0:20000, samples=100] {(3600 + 0.1678*x)/3600};
\addplot[black, dashed] coordinates {(9454,0) (9454,3.3)};
\node[font=\tiny, anchor=south west] at (axis cs:9454,0.1) {break-even $\approx$9.5k queries};
\legend{P1 (no upfront cost), P2 (incl.\ 1h generation)}
\end{axis}
\end{tikzpicture}
\caption{Amortization of the one-time instruction-generation cost. P1 has no upfront cost; P2 includes a worst-case $\approx$1-hour offline generation cost, amortized against per-query savings at $K{=}5$.}
\label{fig:breakeven}
\end{figure}

\begin{table*}[t]
\centering
\begin{tabular}{llcccc}
\toprule
\textbf{Budget} & \textbf{Pipeline}  & \textbf{Tokens} & \textbf{Input TPS} & \textbf{Output TPS} & \textbf{Total TPS} \\
\midrule
\multirow{4}{*}{Low (\textasciitilde1.5k)}
 & P1 (K=1)                     & 1,152 & 2,436.65 & 17.42 & 2,454.07 \\
 & P2 (K=10)        & 1,700 & 6,480.92 & 21.25 & \textbf{6,502.18} \\
 & P3 (K=1, K'=3)                  & 1,493 & 4,165.58 & 18.71 & 4,184.28 \\
 & P4 (K=3, K'=1)             & 1,600 & 3,237.63 & 19.83 & 4,547.96 \\
\midrule
\multirow{4}{*}{Mid (\textasciitilde3.3k)}
 & P1 (K=5)                    & 3,400 & 3,045.66 & 8.20  & 3,053.86 \\
 & P2 (K=25)              & 3,305 & 8,548.72 & 13.28 & \textbf{8,561.99} \\
 & P3 (K=5, K'=3)   & 3,729 & 3,604.90 & 7.58  & 3,612.48 \\
 & P4 (K=10, K'=3)           & 3,360 & 4,121.64 & 8.13  & 4,129.77 \\
\midrule
\multirow{2}{*}{High (\textasciitilde6.3k)}
 & P1 (K=10)       & 6,210 & 2,637.01 & 3.82  & 2,640.82 \\
 & P3 (K=10, K'=3)  & 6,631 & 2,753.74 & 3.54  & \textbf{2,757.28} \\
\bottomrule
\end{tabular}
\caption{Token-level comparison across low, mid, and high token budgets.}
\label{tab:token_budget_comparison}
\end{table*}

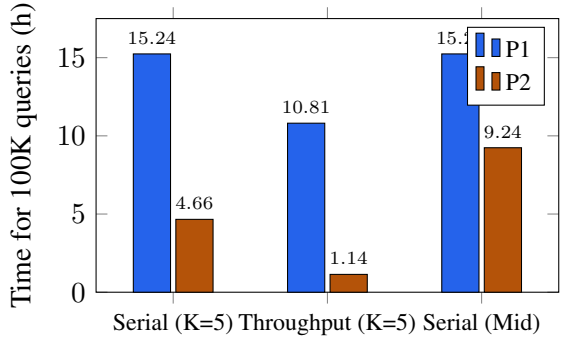
\begin{figure}[t]
\centering
\begin{tikzpicture}
\begin{axis}[
    ybar,
    bar width=14pt,
    width=\columnwidth,
    height=5.2cm,
    symbolic x coords={Serial (K=5), Throughput (K=5), Serial (Mid)},
    xtick=data,
    x tick label style={font=\footnotesize},
    ylabel={Time for 100K queries (h)},
    ymin=0, ymax=17.5,
    enlarge x limits=0.25,
    legend pos=north east,
    legend style={font=\small},
    nodes near coords,
    every node near coord/.append style={font=\scriptsize},
]
\addplot[fill=promptframe-image] coordinates {(Serial (K=5),15.24) (Throughput (K=5),10.81) (Serial (Mid),15.24)};
\addplot[fill=promptframe-instr] coordinates {(Serial (K=5),4.66) (Throughput (K=5),1.14) (Serial (Mid),9.24)};
\legend{P1, P2}
\end{axis}
\end{tikzpicture}
\caption{Projected P1 vs.\ P2 completion time for 100K queries. Reductions: 69.4\% (Serial, $K{=}5$), 89.5\% (Throughput, $K{=}5$), 39.4\% (Serial, token-matched Mid budget). At 1M queries the same ratios scale linearly, extending the savings to $\approx$105.8h (Serial) and $\approx$96.7h (Throughput).}
\label{fig:regime_comparison}
\end{figure}

To contextualize these numbers at deployment scale, we project cumulative time for 100,000 queries under two serving regimes: serial (single-query latency, no concurrency) and throughput-based (matched to our benchmarked QPS, reflecting realistic concurrent serving), shown in Figure~\ref{fig:regime_comparison}. At $K{=}5$, the serial regime saves 69.4\% of wall-clock time and the throughput regime saves 89.5\%; isolating the modality effect at the token-matched Mid budget (P1 $K{=}5$ vs.\ P2 $K{=}25$) still yields a 39.4\% serial-regime saving, confirming that the reduction is not solely a token-count artifact.

We caution that the throughput-regime figures assume our benchmarked QPS holds at scale, which depends on GPU memory and batching configuration; real production savings will vary with serving infrastructure, but the direction and approximate magnitude of the effect, a 3-9x improvement in effective query capacity for instruction ICL over image ICL at matched K, should hold under similar serving conditions.

Unlike P1, P2 carries a one-time upfront cost: instructions must be generated offline for every training image before any query is served (Appendix~\ref{sec:appendix_impl} reports this took minutes to, at most, an hour per dataset). Figure~\ref{fig:breakeven} amortizes this worst-case cost against the $K{=}5$ serial-regime per-query latencies (Table~\ref{tab:pipeline_comparison}): the two curves cross at $\approx$9,500 queries, after which P2 is strictly cheaper and the gap widens with scale. Since generation typically takes minutes rather than the reported maximum of one hour, this is a conservative estimate and the true break-even point is usually earlier.
\section{Token Cost Derivation}
\label{sec:appendix_tokens}

Both models in our evaluation, Gemma-3-4B-IT and Qwen2.5-VL-7B-Instruct, are served using vLLM, with each model's native HuggingFace processor handling image preprocessing and tokenization. Token counts reported in Table \ref{tab:token_components} are measured directly from the tokenized prompts produced by these processors, on average.
With the CLS (class) token, this yields 577 vision tokens per image.
In practice, after accounting for the image special tokens injected by
each model's chat template, the effective per-image token count is
$\approx$550 (measured empirically by inspecting tokenizer output across
all four models). 

Table \ref{tab:token_components} reports the average token cost of each component that appears in our prompts, measured from the vLLM-served tokenizer output. System prompts contribute between 15 and 40 tokens. A single image example consumes approximately 550 tokens for the patch embeddings plus approximately 12 tokens for its label text. For P1(b), each in-context image is compressed to approximately 224$\times$152 at 75\% JPEG quality before being tokenized, reducing the per-image patch cost to approximately 258 tokens; the query image is left at full resolution in every pipeline, so this reduction applies only to the retrieved context images. A single distilled instruction (three rules) costs approximately 95 tokens, plus approximately 12 tokens for the associated label. The query image itself contributes approximately 550 tokens, the target/classification prompt contributes between 24 and 80 tokens, and the model output (the predicted class name) contributes between 2 and 5 tokens. Instruction token counts are measured as the mean across all generated instructions.

\begin{table}[h]
\centering\small
\begin{tabular}{lr}
\toprule
\textbf{Component} & \textbf{Tokens (avg)} \\
\midrule
System prompt & 15--40 \\
Per image example (patches) & $\approx$550 \\
Per compressed image example (patches) & $\approx$258 \\
Per image example (label text) & $\approx$12 \\
Per instruction (3 rules) & $\approx$95 \\
Per instruction (label text) & $\approx$12 \\
Query image & $\approx$550 \\
Target / classification prompt & 24--80 \\
Model output (class name) & 2--5 \\
\bottomrule
\end{tabular}
\caption{\textbf{Per-component token counts}.
  Instruction tokens are measured as the mean token count of generated
  instructions across all training images and all models.}
\label{tab:token_components}
\end{table}

\paragraph{Per-Pipeline Token Budget.} Summing the components in Table~\ref{tab:token_components} gives per-pipeline cost:
\begin{align}
  &C_{\text{P0}} \!=\! 550 \!+\! 24 \!=\! 574 \nonumber \\
  &C_{\text{P1(a)}}(K) \!=\! K \!\times\! (550 + 12) \!+\! 550 \!+\! 40 \nonumber \\
  &C_{\text{P1(b)}}(K) \!=\! K \!\times\! (258 + 12) \!+\! 550 \!+\! 40 \nonumber \\
  &C_{\text{P2}}(K) \!=\! K \!\times\! (95 + 12) + 550 + 80 \nonumber \\
  &C_{\text{P3}}(K,K') \!=\! K \!\times\! (550 + 9) \!+\! K' \!\times\! (95 \!+\! 12) \!+\! 550 \!+\! 170 \nonumber \\
  &C_{\text{P4}}(K,K') \!=\! K \!\times\! (95\! + \!12) + K' \!\times\! (550 \!+\! 9) \!+\! 550 \!+\! 170 \nonumber
\end{align}
$C_{\text{P1(b)}}$ follows the same form as $C_{\text{P1(a)}}$ with the per-image patch cost replaced by the compressed value (258 vs.\ 550 tokens); the query image, system prompt, and target prompt are unaffected since compression is applied only to the retrieved in-context images. At $K{=}5$: $C_{\text{P1(a)}} \approx 3{,}390$, $C_{\text{P1(b)}} \approx 1{,}940$, and $C_{\text{P2}} \approx
1{,}165$, yielding a P1(a)/P2 ratio of $\mathbf{2.91\times}$ and a P1(b)/P2 ratio of $\mathbf{1.67\times}$.

\clearpage
\onecolumn
\section{Generated Instruction Examples}
\label{sec:appendix_instructions}
We show example identification instructions generated by two MLLMs for three
training images from each dataset mentioned in Appendix \ref{sec:appendix_datasets}. Each instruction consists of three
visual reasoning rules generated from the image and its class label.

\vspace{4pt}
\noindent\textbf{Legend:}\quad
\colorbox{promptbg-instr}{\strut\hspace{4pt}\textcolor{promptframe-instr}{\rule{8pt}{8pt}}\hspace{4pt}}
\hspace{2pt} Gemma-3-4B-IT \quad
\colorbox{promptbg-image}{\strut\hspace{4pt}\textcolor{promptframe-image}{\rule{8pt}{8pt}}\hspace{4pt}}
\hspace{2pt} Qwen2.5-VL-7B-Instruct
\vspace{6pt}
\tcbset{
  gemmabox/.style={enhanced, left=5pt, right=5pt, top=4pt, bottom=4pt,
    fontupper=\small, boxrule=1.2pt, arc=0pt, width=\linewidth,
    colback=promptbg-instr, colframe=promptframe-instr},
  qwenbox/.style={enhanced, left=5pt, right=5pt, top=4pt, bottom=4pt,
    fontupper=\small, boxrule=1.2pt, arc=0pt, width=\linewidth,
    colback=promptbg-image, colframe=promptframe-image}
}

\subsection*{Food-101}

\noindent
\begin{minipage}[t]{0.32\textwidth}
  \centering
  \includegraphics[width=\linewidth,height=5cm,keepaspectratio]{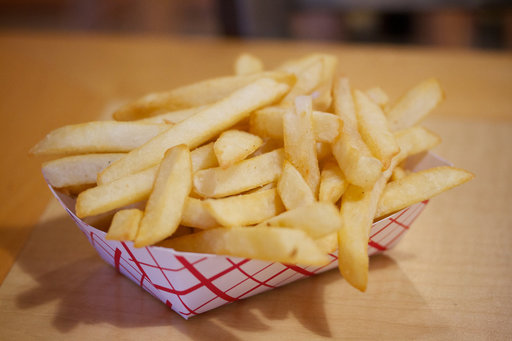}\\[4pt]
  {\small\textit{Ground Truth: \textbf{french fries}}}
  \vspace{8pt}

  \begin{tcolorbox}[gemmabox]
    \begin{enumerate}[leftmargin=*,topsep=0pt,itemsep=2pt,parsep=0pt]
      \item To identify french fries, look for elongated, irregular-shaped potato strips arranged in a pile.
      \item To distinguish from similar dishes, focus on a pale yellow, slightly browned exterior and a fluffy, airy interior texture.
      \item To avoid misclassification, check for a paper container with a red and white checkered pattern.
    \end{enumerate}
  \end{tcolorbox}
  \vspace{4pt}
  \begin{tcolorbox}[qwenbox]
    \begin{enumerate}[leftmargin=*,topsep=0pt,itemsep=2pt,parsep=0pt]
      \item To identify french fries, look for long, thin, elongated pieces of golden-brown fried potato.
      \item To distinguish from similar dishes, focus on the crispy, slightly uneven surface texture and the absence of visible garnishes or sauces.
      \item To avoid misclassification, check for the presence of other ingredients like cheese or vegetables, which would indicate a different dish.
    \end{enumerate}
  \end{tcolorbox}
\end{minipage}
\hfill
\begin{minipage}[t]{0.32\textwidth}
  \centering
  \includegraphics[width=\linewidth,height=5cm,keepaspectratio]{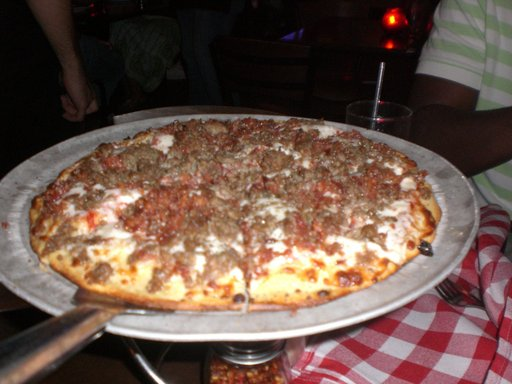}\\[4pt]
  {\small\textit{Ground Truth: \textbf{pizza}}}
  \vspace{8pt}

  \begin{tcolorbox}[gemmabox]
    \begin{enumerate}[leftmargin=*,topsep=0pt,itemsep=2pt,parsep=0pt]
      \item To identify pizza, look for a round, flat, baked dough base covered with toppings.
      \item To distinguish from similar dishes, focus on the presence of melted cheese and visible layers of sauce, toppings, and crust.
      \item To avoid misclassification, check for a browned, bubbly crust and a variety of toppings like meat, vegetables, and cheese.
    \end{enumerate}
  \end{tcolorbox}
  \vspace{4pt}
  \begin{tcolorbox}[qwenbox]
    \begin{enumerate}[leftmargin=*,topsep=0pt,itemsep=2pt,parsep=0pt]
      \item To identify pizza, look for a round, flatbread base topped with tomato sauce, cheese, and various toppings like meats, vegetables, or herbs.
      \item To distinguish from similar dishes, focus on the thin, crispy crust and the combination of melted cheese and toppings spread evenly across the surface.
      \item To avoid misclassification, check for the absence of a thick, bread-like texture and the presence of a single continuous layer of toppings.
    \end{enumerate}
  \end{tcolorbox}
\end{minipage}
\hfill
\begin{minipage}[t]{0.32\textwidth}
  \centering
  \includegraphics[width=\linewidth,height=5cm,keepaspectratio]{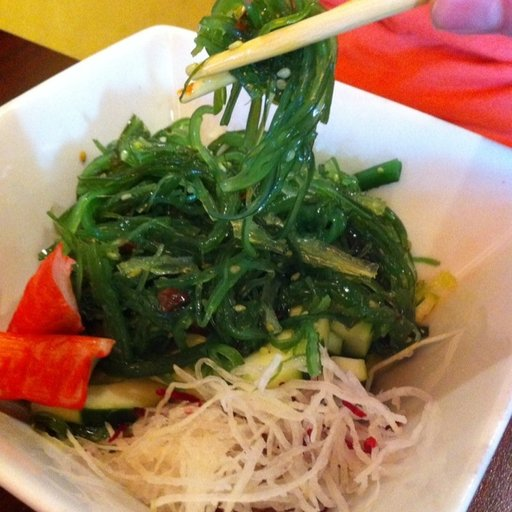}\\[4pt]
  {\small\textit{Ground Truth: \textbf{seaweed salad}}}
  \vspace{8pt}

  \begin{tcolorbox}[gemmabox]
    \begin{enumerate}[leftmargin=*,topsep=0pt,itemsep=2pt,parsep=0pt]
      \item To identify seaweed salad, look for a bowl filled with dark green, wavy seaweed as the dominant ingredient.
      \item To distinguish from similar salads, focus on the presence of sesame seeds and a light, often citrus-based dressing drizzled over the seaweed.
      \item To avoid misclassification, check that there are no visible chunks of meat or other substantial proteins, differentiating it from a seafood salad.
    \end{enumerate}
  \end{tcolorbox}
  \vspace{4pt}
  \begin{tcolorbox}[qwenbox]
    \begin{enumerate}[leftmargin=*,topsep=0pt,itemsep=2pt,parsep=0pt]
      \item To identify seaweed salad, look for green, stringy seaweed mixed with other vegetables like cucumber and shredded radish.
      \item To distinguish from similar dishes, focus on the presence of a light dressing or sauce coating the seaweed and vegetables.
      \item To avoid misclassification, check for the absence of browning or char marks, which would indicate a different cooking method.
    \end{enumerate}
  \end{tcolorbox}
\end{minipage}

\newpage

\subsection*{EuroSAT}

\noindent
\begin{minipage}[t]{0.32\textwidth}
  \centering
  \includegraphics[width=\linewidth,height=5cm,keepaspectratio]{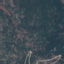}\\[4pt]
  {\small\textit{Ground Truth: \textbf{herbaceous vegetation}}}
  \vspace{8pt}

  \begin{tcolorbox}[gemmabox]
    \begin{enumerate}[leftmargin=*,topsep=0pt,itemsep=2pt,parsep=0pt]
      \item To identify herbaceous vegetation, look for areas exhibiting a dominant green to yellow-green color tone across the visible spectrum.
      \item To distinguish from similar classes like shrubs or forests, focus on the relatively uniform texture and lack of distinct, woody outlines.
      \item To avoid misclassification, check for the absence of significant shadows or snow cover, which can obscure the characteristic green reflectance.
    \end{enumerate}
  \end{tcolorbox}
  \vspace{4pt}
  \begin{tcolorbox}[qwenbox]
    \begin{enumerate}[leftmargin=*,topsep=0pt,itemsep=2pt,parsep=0pt]
      \item To identify herbaceous vegetation, look for light green to yellowish areas with fine texture.
      \item To distinguish from similar classes, focus on the uniformity in color and texture, and the absence of distinct structural elements like trees or buildings.
      \item To avoid misclassification, check for the presence of bare soil patches nearby to confirm the vegetation is indeed herbaceous.
    \end{enumerate}
  \end{tcolorbox}
\end{minipage}
\hfill
\begin{minipage}[t]{0.32\textwidth}
  \centering
  \includegraphics[width=\linewidth,height=5cm,keepaspectratio]{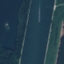}\\[4pt]
  {\small\textit{Ground Truth: \textbf{river}}}
  \vspace{8pt}

  \begin{tcolorbox}[gemmabox]
    \begin{enumerate}[leftmargin=*,topsep=0pt,itemsep=2pt,parsep=0pt]
      \item To identify the river, look for long, linear features with a distinct, darker tone compared to surrounding vegetation.
      \item To distinguish from similar classes like lakes or flooded areas, focus on the presence of a defined channel and sediment deposition along the banks.
      \item To avoid misclassification, check for the absence of significant vegetation growth directly adjacent to the channel, indicating a consistently flowing water body.
    \end{enumerate}
  \end{tcolorbox}
  \vspace{4pt}
  \begin{tcolorbox}[qwenbox]
    \begin{enumerate}[leftmargin=*,topsep=0pt,itemsep=2pt,parsep=0pt]
      \item To identify the river, look for a linear feature with a consistent width and a distinct darker tone compared to surrounding areas.
      \item To distinguish from similar classes, focus on reflections or shadows typical of water bodies, which are absent in land uses like roads or trails.
      \item To avoid misclassification, check for the absence of vegetation along the banks, which is common in rivers but rare in other linear features.
    \end{enumerate}
  \end{tcolorbox}
\end{minipage}
\hfill
\begin{minipage}[t]{0.32\textwidth}
  \centering
  \includegraphics[width=\linewidth,height=5cm,keepaspectratio]{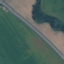}\\[4pt]
  {\small\textit{Ground Truth: \textbf{highway}}}
  \vspace{8pt}

  \begin{tcolorbox}[gemmabox]
    \begin{enumerate}[leftmargin=*,topsep=0pt,itemsep=2pt,parsep=0pt]
      \item To identify the highway, look for a long, linear feature with a relatively uniform gray or light-colored surface.
      \item To distinguish from similar classes like roads, focus on the consistent, straight alignment and lack of significant branching or intersections.
      \item To avoid misclassification, check for the presence of guardrails or signage, which are often associated with highways but less common on minor roads.
    \end{enumerate}
  \end{tcolorbox}
  \vspace{4pt}
  \begin{tcolorbox}[qwenbox]
    \begin{enumerate}[leftmargin=*,topsep=0pt,itemsep=2pt,parsep=0pt]
      \item To identify the highway, look for long, straight linear features with a consistent width.
      \item To distinguish from similar classes, focus on the presence of vehicles, road markings, and surrounding infrastructure like bridges or overpasses.
      \item To avoid misclassification, check for uniformity in the road surface texture and compare with known highway characteristics in other images.
    \end{enumerate}
  \end{tcolorbox}
\end{minipage}

\newpage

\subsection*{Flowers102}

\noindent
\begin{minipage}[t]{0.32\textwidth}
  \centering
  \includegraphics[width=\linewidth,height=5cm,keepaspectratio]{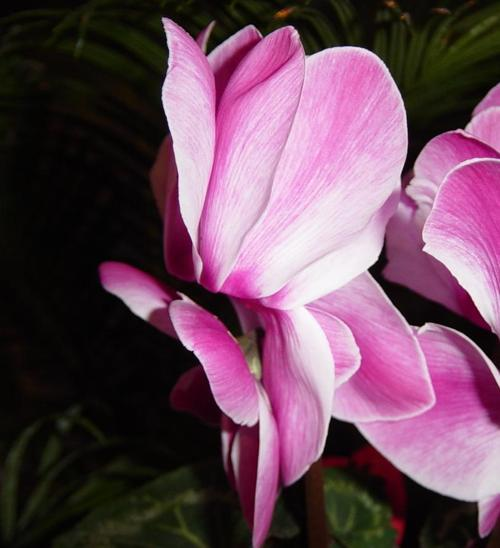}\\[4pt]
  {\small\textit{Ground Truth: \textbf{cyclamen}}}
  \vspace{8pt}

  \begin{tcolorbox}[gemmabox]
    \begin{enumerate}[leftmargin=*,topsep=0pt,itemsep=2pt,parsep=0pt]
      \item To identify cyclamen, look for five petals arranged in radial symmetry, with rounded petals and a gradient color pattern transitioning from white to pink at the edges.
      \item To distinguish from visually similar species, focus on the presence of distinct eye zones within the center of the flower petals.
      \item To avoid misclassification, differentiate from \textit{Cypripedium} by checking that the petals are attached directly to the stem, not to a broad shield-shaped base.
    \end{enumerate}
  \end{tcolorbox}
  \vspace{4pt}
  \begin{tcolorbox}[qwenbox]
    \begin{enumerate}[leftmargin=*,topsep=0pt,itemsep=2pt,parsep=0pt]
      \item To identify cyclamen, look for five petals with a rounded shape, radially symmetric bloom, and bicolor center with pink and white tones.
      \item To distinguish from visually similar species, focus on the smooth petal texture and the presence of prominent stamens.
      \item To avoid misclassification, check for the distinctive eye zone at the center of the petals, which distinguishes cyclamen from other similar species like violets.
    \end{enumerate}
  \end{tcolorbox}
\end{minipage}
\hfill
\begin{minipage}[t]{0.32\textwidth}
  \centering
  \includegraphics[width=\linewidth,height=5cm,keepaspectratio]{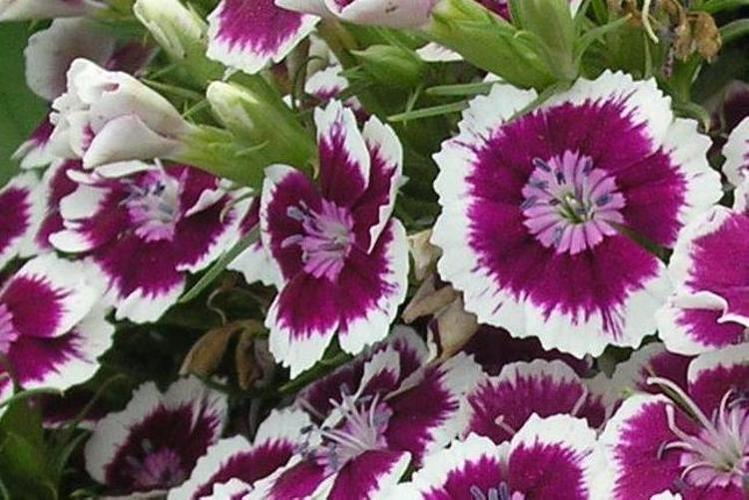}\\[4pt]
  {\small\textit{Ground Truth: \textbf{sweet william}}}
  \vspace{8pt}

  \begin{tcolorbox}[gemmabox]
    \begin{enumerate}[leftmargin=*,topsep=0pt,itemsep=2pt,parsep=0pt]
      \item To identify sweet william, look for radially symmetric blooms with six petals exhibiting a bicolor center pattern of pink and white.
      \item To distinguish from visually similar species, focus on the prominent, contrasting yellow stamen arrangement within the flower's center.
      \item To avoid misclassification, differentiate from dianthus and carnations by the sweet william's six petals versus dianthus's five and the carnation's ruffled petals.
    \end{enumerate}
  \end{tcolorbox}
  \vspace{4pt}
  \begin{tcolorbox}[qwenbox]
    \begin{enumerate}[leftmargin=*,topsep=0pt,itemsep=2pt,parsep=0pt]
      \item To identify sweet william, look for flowers with five petals, rounded to slightly fringed petals, radially symmetric blooms, and a bicolor center with a prominent eye zone.
      \item To distinguish from visually similar species, focus on the distinct eye zone with a dark center surrounded by a white ring.
      \item To avoid misclassification, check for the prominent eye zone with a dark center and white ring, distinguishing it from other bicolor flowers like dianthus.
    \end{enumerate}
  \end{tcolorbox}
\end{minipage}
\hfill
\begin{minipage}[t]{0.32\textwidth}
  \centering
  \includegraphics[width=\linewidth,height=5cm,keepaspectratio]{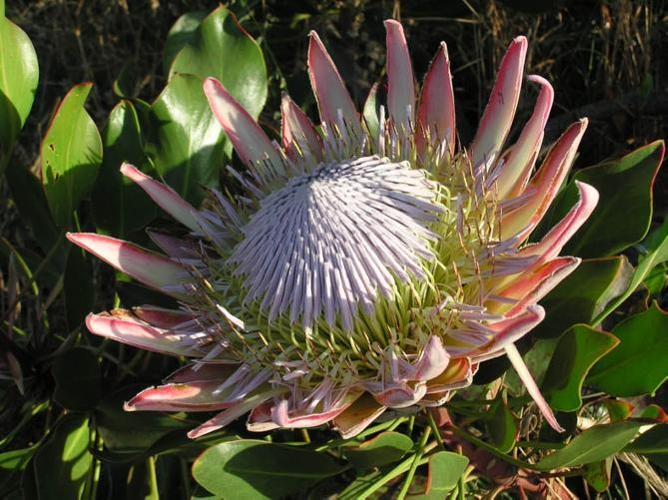}\\[4pt]
  {\small\textit{Ground Truth: \textbf{king protea}}}
  \vspace{8pt}

  \begin{tcolorbox}[gemmabox]
    \begin{enumerate}[leftmargin=*,topsep=0pt,itemsep=2pt,parsep=0pt]
      \item To identify king protea, look for a radially symmetric bloom with pointed petals exhibiting a bicolor center of pink and cream.
      \item To distinguish from visually similar species, focus on the prominent, recessed stamen arrangement within the central disc.
      \item To avoid misclassification, differentiate from \textit{Protea magnifica} by the king protea's distinctly elongated, pointed petals compared to the \textit{magnifica}'s more rounded ones.
    \end{enumerate}
  \end{tcolorbox}
  \vspace{4pt}
  \begin{tcolorbox}[qwenbox]
    \begin{enumerate}[leftmargin=*,topsep=0pt,itemsep=2pt,parsep=0pt]
      \item To identify king protea, look for a large, radially symmetric bloom with numerous pointed, reflexed petals surrounding a prominent, densely packed central cone.
      \item To distinguish from visually similar species, focus on the distinctively textured, velvety surface of the petals and the prominent central cone structure.
      \item To avoid misclassification, check for the presence of a deep red eye zone at the base of the petals, which is characteristic of the king protea.
    \end{enumerate}
  \end{tcolorbox}
\end{minipage}

\newpage

\subsection*{SUN397}

\noindent
\begin{minipage}[t]{0.32\textwidth}
  \centering
  \includegraphics[width=\linewidth,height=5cm,keepaspectratio]{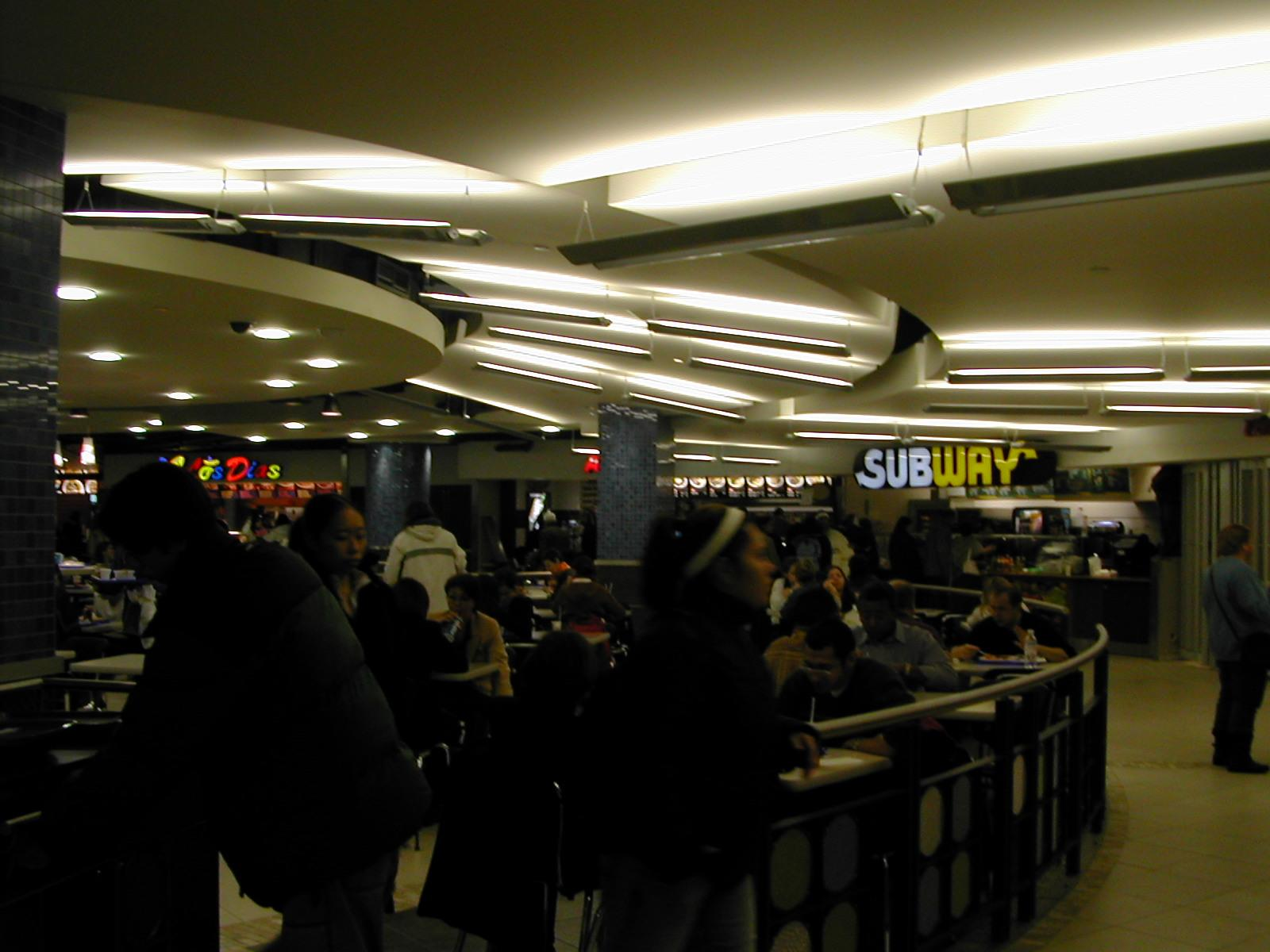}\\[4pt]
  {\small\textit{Ground Truth: \textbf{food court}}}
  \vspace{8pt}

  \begin{tcolorbox}[gemmabox]
    \begin{enumerate}[leftmargin=*,topsep=0pt,itemsep=2pt,parsep=0pt]
      \item To identify food court, look for enclosed indoor spaces with a high density of seating areas, food service counters, and retail outlets with artificial lighting.
      \item To distinguish from similar indoor spaces, focus on the presence of multiple food service vendors and seating designed for immediate consumption.
      \item To avoid misclassification, check for signage advertising food and beverage options, differentiating it from a transit station or shopping mall corridor.
    \end{enumerate}
  \end{tcolorbox}
  \vspace{4pt}
  \begin{tcolorbox}[qwenbox]
    \begin{enumerate}[leftmargin=*,topsep=0pt,itemsep=2pt,parsep=0pt]
      \item To identify food court, look for an indoor, expansive environment with a combination of seating areas, food stalls, and signage indicating food vendors.
      \item To distinguish from visually similar scenes, focus on multiple food stalls with distinct signage, communal seating, and food-related items such as condiments and serving trays.
      \item To avoid misclassification, check for the absence of medical equipment, educational furniture, or industrial machinery.
    \end{enumerate}
  \end{tcolorbox}
\end{minipage}
\hfill
\begin{minipage}[t]{0.32\textwidth}
  \centering
  \includegraphics[width=\linewidth,height=5cm,keepaspectratio]{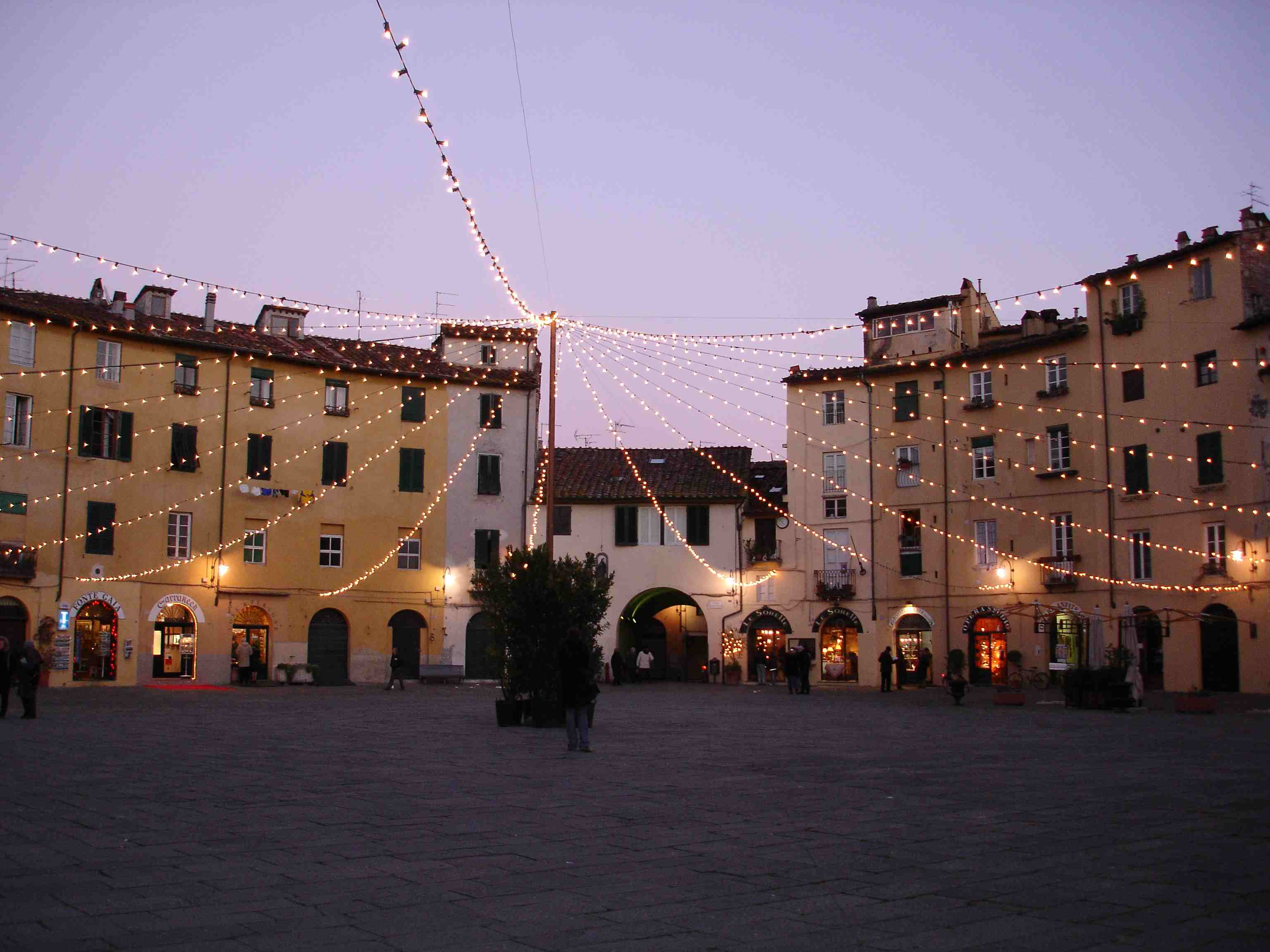}\\[4pt]
  {\small\textit{Ground Truth: \textbf{plaza}}}
  \vspace{8pt}

  \begin{tcolorbox}[gemmabox]
    \begin{enumerate}[leftmargin=*,topsep=0pt,itemsep=2pt,parsep=0pt]
      \item To identify plaza, look for a paved open space with buildings surrounding it, featuring storefronts and often adorned with string lights and decorative fixtures.
      \item To distinguish from similar urban squares, focus on multiple storefronts with visible signage and consistent stone or brick paving throughout.
      \item To avoid misclassification, check for a central fountain or statue, which distinguishes a plaza from a simple street intersection or courtyard.
    \end{enumerate}
  \end{tcolorbox}
  \vspace{4pt}
  \begin{tcolorbox}[qwenbox]
    \begin{enumerate}[leftmargin=*,topsep=0pt,itemsep=2pt,parsep=0pt]
      \item To identify plaza, look for an open, expansive outdoor space with buildings forming a perimeter, typically featuring paved ground and festive elements like string lights.
      \item To distinguish from visually similar scenes, focus on benches, fountains, or market stalls that anchor its function as a public gathering area.
      \item To avoid misclassification, check for architectural elements such as arches, columns, or statues common in plazas but less so in other open spaces.
    \end{enumerate}
  \end{tcolorbox}
\end{minipage}
\hfill
\begin{minipage}[t]{0.32\textwidth}
  \centering
  \includegraphics[width=\linewidth,height=5cm,keepaspectratio]{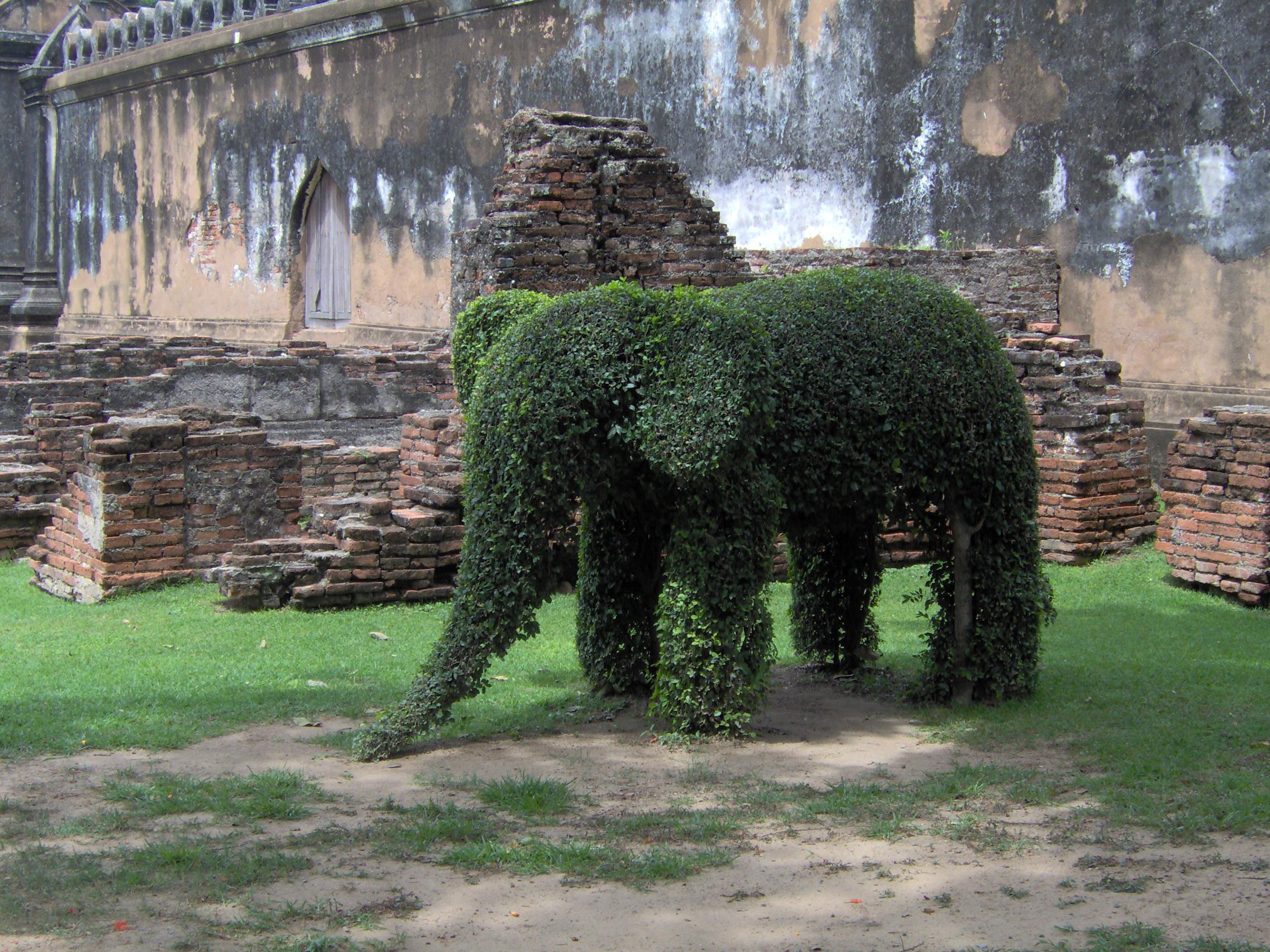}\\[4pt]
  {\small\textit{Ground Truth: \textbf{topiary garden}}}
  \vspace{8pt}

  \begin{tcolorbox}[gemmabox]
    \begin{enumerate}[leftmargin=*,topsep=0pt,itemsep=2pt,parsep=0pt]
      \item To identify topiary garden, look for dense, meticulously shaped foliage in geometric patterns within a landscaped space with defined pathways.
      \item To distinguish from similar outdoor spaces, focus on the prevalence of sculpted shrubs and hedges indicating deliberate horticultural design.
      \item To avoid misclassification, check for classical garden features like stone lanterns or statuary, which differentiate it from more naturalistic landscapes.
    \end{enumerate}
  \end{tcolorbox}
  \vspace{4pt}
  \begin{tcolorbox}[qwenbox]
    \begin{enumerate}[leftmargin=*,topsep=0pt,itemsep=2pt,parsep=0pt]
      \item To identify topiary garden, look for an outdoor setting with large, sculpted topiary plants, surrounded by other ornamental greenery and garden elements.
      \item To distinguish from visually similar scenes, focus on multiple topiary plants in a formal layout, featuring hedges and ornamental shrubs.
      \item To avoid misclassification, check for the absence of indoor elements and the presence of natural sunlight and outdoor gardening equipment.
    \end{enumerate}
  \end{tcolorbox}
\end{minipage}

\newpage
\subsection*{Describable Textures (DTD)}

\noindent
\begin{minipage}[t]{0.32\textwidth}
  \centering
  \includegraphics[width=\linewidth,height=5cm,keepaspectratio]{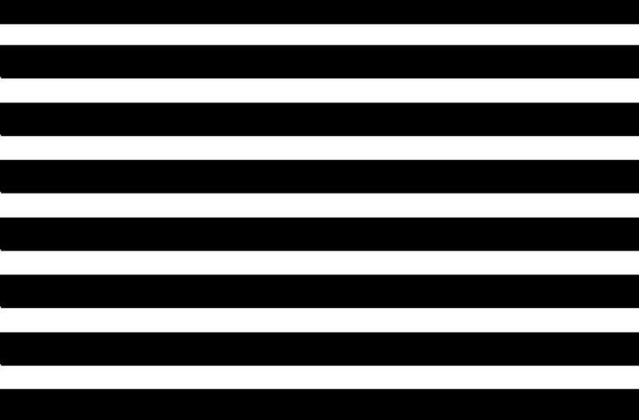}\\[4pt]
  {\small\textit{Ground Truth: \textbf{banded}}}
  \vspace{8pt}

  \begin{tcolorbox}[gemmabox]
    \begin{enumerate}[leftmargin=*,topsep=0pt,itemsep=2pt,parsep=0pt]
      \item To identify the banded texture, look for a consistent, repeating pattern of parallel lines with varying widths and spacing.
      \item To distinguish from similar textures, focus on the regular, linear arrangement of these bands rather than a random or chaotic distribution.
      \item To avoid misclassification, check for a clear sense of directionality, indicating a dominant orientation within the image.
    \end{enumerate}
  \end{tcolorbox}
  \vspace{4pt}
  \begin{tcolorbox}[qwenbox]
    \begin{enumerate}[leftmargin=*,topsep=0pt,itemsep=2pt,parsep=0pt]
      \item To identify the banded texture, look for horizontal stripes of equal width and consistent spacing across the entire image.
      \item To distinguish from similar textures, focus on the uniformity of stripe width and spacing and the absence of irregularities.
      \item To avoid misclassification, check for a consistent, regular, evenly spaced stripe pattern without any breaks or interruptions.
    \end{enumerate}
  \end{tcolorbox}
\end{minipage}
\hfill
\begin{minipage}[t]{0.32\textwidth}
  \centering
  \includegraphics[width=\linewidth,height=5cm,keepaspectratio]{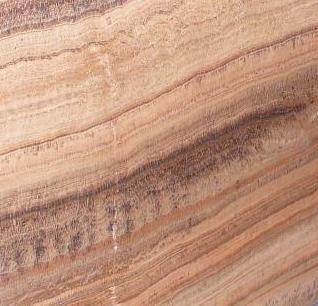}\\[4pt]
  {\small\textit{Ground Truth: \textbf{marbled}}}
  \vspace{8pt}

  \begin{tcolorbox}[gemmabox]
    \begin{enumerate}[leftmargin=*,topsep=0pt,itemsep=2pt,parsep=0pt]
      \item To identify the marbled texture, look for parallel, undulating bands of varying color and tone that create a layered appearance.
      \item To distinguish from similar textures, focus on the consistent, repeating pattern of bands rather than random variation in color or form.
      \item To avoid misclassification, check for a sense of depth or layering where bands appear to overlap and recede into the distance.
    \end{enumerate}
  \end{tcolorbox}
  \vspace{4pt}
  \begin{tcolorbox}[qwenbox]
    \begin{enumerate}[leftmargin=*,topsep=0pt,itemsep=2pt,parsep=0pt]
      \item To identify the marbled texture, look for irregular patterns of light and dark bands that curve and twist across the surface.
      \item To distinguish from similar textures, focus on distinct, non-uniform banding rather than smooth gradients or uniform speckling.
      \item To avoid misclassification, check for the absence of consistent parallel lines or uniform coloration that would suggest wood grain or stone.
    \end{enumerate}
  \end{tcolorbox}
\end{minipage}
\hfill
\begin{minipage}[t]{0.32\textwidth}
  \centering
  \includegraphics[width=\linewidth,height=5cm,keepaspectratio]{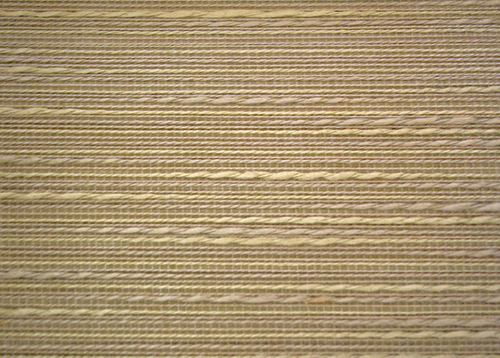}\\[4pt]
  {\small\textit{Ground Truth: \textbf{woven}}}
  \vspace{8pt}

  \begin{tcolorbox}[gemmabox]
    \begin{enumerate}[leftmargin=*,topsep=0pt,itemsep=2pt,parsep=0pt]
      \item To identify the woven texture, look for a repeating, interlaced pattern of parallel and perpendicular lines formed by multiple distinct threads.
      \item To distinguish from similar textures, focus on the clear, regular spacing between individual threads creating a grid-like structure.
      \item To avoid misclassification, check for surface relief caused by raised or recessed threads, indicating three-dimensional construction of the weave.
    \end{enumerate}
  \end{tcolorbox}
  \vspace{4pt}
  \begin{tcolorbox}[qwenbox]
    \begin{enumerate}[leftmargin=*,topsep=0pt,itemsep=2pt,parsep=0pt]
      \item To identify the woven texture, look for a repeating pattern of intersecting lines forming a grid-like structure.
      \item To distinguish from similar textures, focus on the regularity and uniformity of the line spacing and direction.
      \item To avoid misclassification, check for consistent, parallel lines running in two perpendicular directions.
    \end{enumerate}
  \end{tcolorbox}
\end{minipage}

\newpage
\subsection*{FGVC Aircraft}

\noindent
\begin{minipage}[t]{0.32\textwidth}
  \centering
  \includegraphics[width=\linewidth,height=5cm,keepaspectratio]{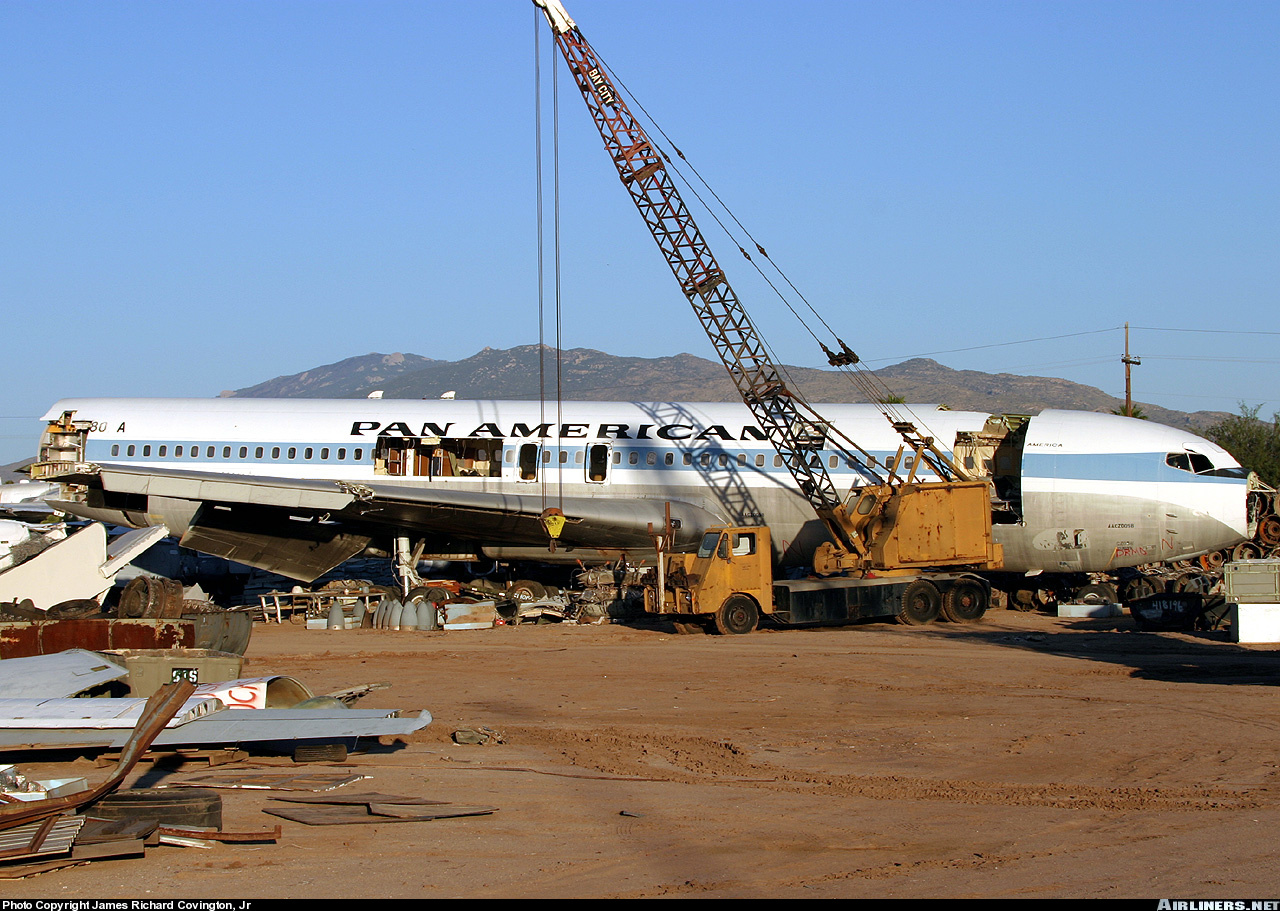}\\[4pt]
  {\small\textit{Ground Truth: \textbf{707-320}}}
  \vspace{8pt}

  \begin{tcolorbox}[gemmabox]
    \begin{enumerate}[leftmargin=*,topsep=0pt,itemsep=2pt,parsep=0pt]
      \item To identify the 707-320, look for a fuselage with a distinctive configuration and a relatively short length compared to later 707 variants.
      \item To distinguish from similar variants, focus on the under-wing engine pylons and the subtle taper of the tail cone.
      \item To avoid misclassification, check for the absence of winglets, a key differentiator from later 707 models with blended winglets.
    \end{enumerate}
  \end{tcolorbox}
  \vspace{4pt}
  \begin{tcolorbox}[qwenbox]
    \begin{enumerate}[leftmargin=*,topsep=0pt,itemsep=2pt,parsep=0pt]
      \item To identify the 707-320, look for four under-wing pylon-mounted engines, a T-tail configuration, and a pointed nose profile.
      \item To distinguish from similar variants, focus on the fuselage length measured by the number of windows between front and rear exits.
      \item To avoid misclassification, check for the absence of winglets; the 707-320 does not have winglets.
    \end{enumerate}
  \end{tcolorbox}
\end{minipage}
\hfill
\begin{minipage}[t]{0.32\textwidth}
  \centering
  \includegraphics[width=\linewidth,height=5cm,keepaspectratio]{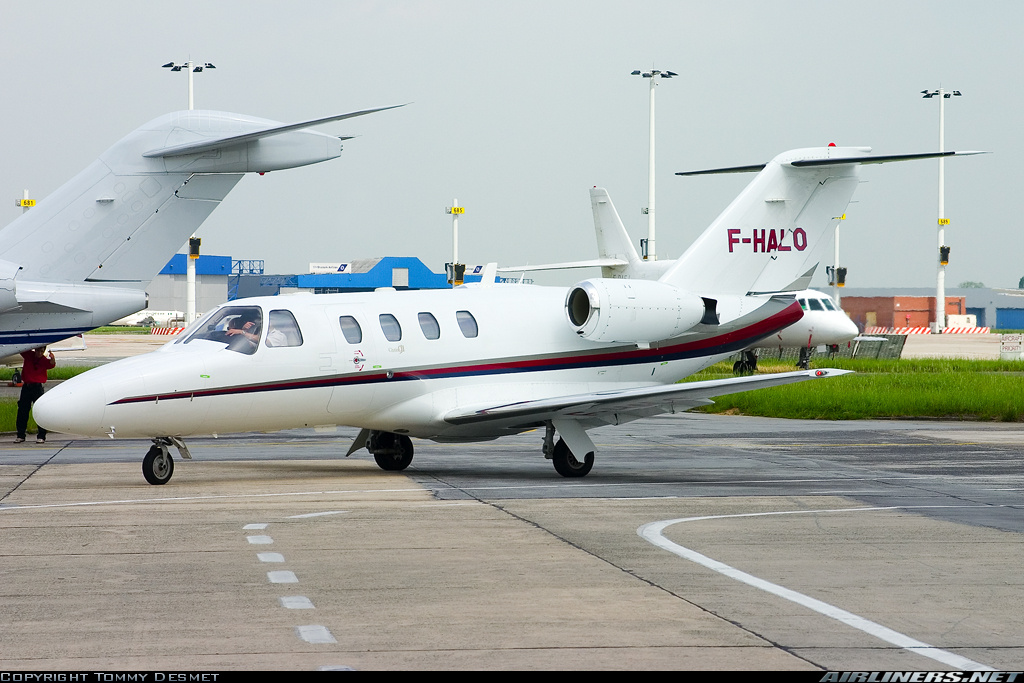}\\[4pt]
  {\small\textit{Ground Truth: \textbf{Cessna 525}}}
  \vspace{8pt}

  \begin{tcolorbox}[gemmabox]
    \begin{enumerate}[leftmargin=*,topsep=0pt,itemsep=2pt,parsep=0pt]
      \item To identify the Cessna 525, look for a twin-engine configuration with under-wing pylons featuring flat-bottomed engine nacelles.
      \item To distinguish from similar variants, focus on the fuselage length indicated by the three evenly spaced doors and window count.
      \item To avoid misclassification, check for the absence of winglets, differentiating it from the Cessna 560 Sovereign models.
    \end{enumerate}
  \end{tcolorbox}
  \vspace{4pt}
  \begin{tcolorbox}[qwenbox]
    \begin{enumerate}[leftmargin=*,topsep=0pt,itemsep=2pt,parsep=0pt]
      \item To identify the Cessna 525, look for two under-wing pylon-mounted engines with circular nacelles, a narrow fuselage, swept-back wings, and a T-tail.
      \item To distinguish from similar variants, focus on the fuselage length and number of windows between front and rear doors.
      \item To avoid misclassification, check the winglet style; the Cessna 525 has blended winglets, while rivals like the Hawker 4000 have sharklets.
    \end{enumerate}
  \end{tcolorbox}
\end{minipage}
\hfill
\begin{minipage}[t]{0.32\textwidth}
  \centering
  \includegraphics[width=\linewidth,height=5cm,keepaspectratio]{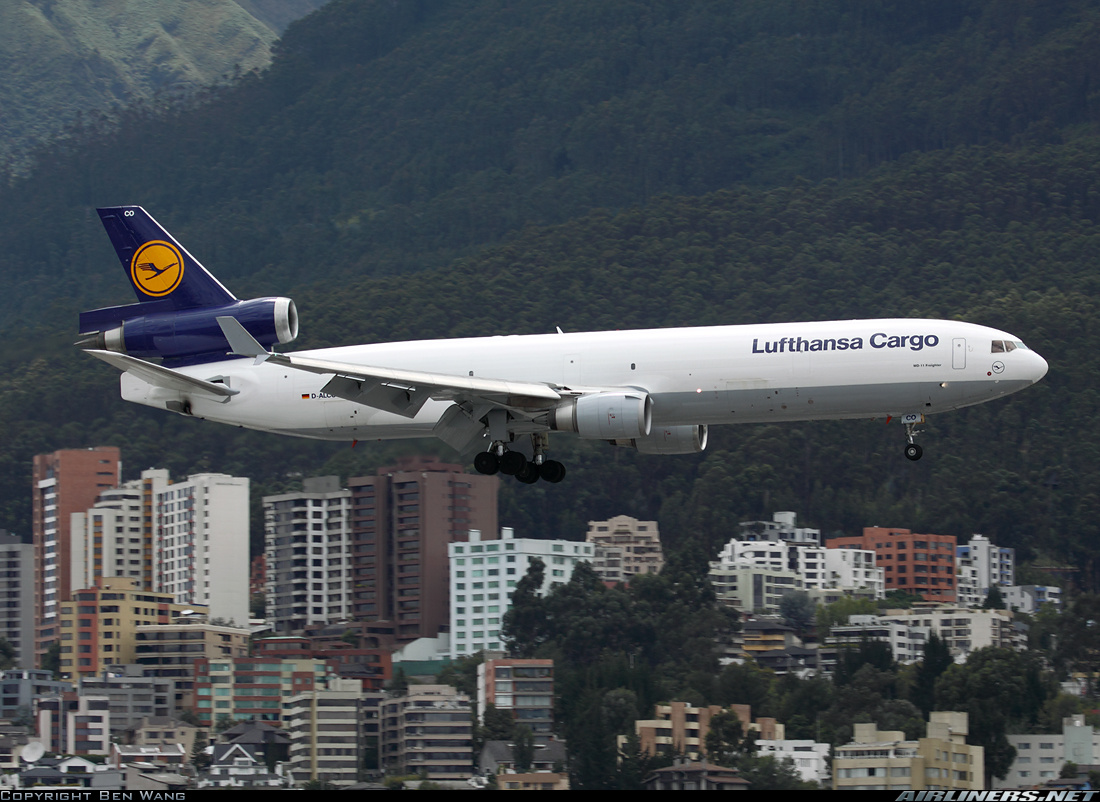}\\[4pt]
  {\small\textit{Ground Truth: \textbf{MD-11}}}
  \vspace{8pt}

  \begin{tcolorbox}[gemmabox]
    \begin{enumerate}[leftmargin=*,topsep=0pt,itemsep=2pt,parsep=0pt]
      \item To identify the MD-11, look for three under-wing engine pylons and a relatively long, blunt fuselage with a pronounced drooped nose.
      \item To distinguish from similar variants, focus on the distinctive blended winglet configuration on the wingtips.
      \item To avoid misclassification, check for the prominent, tapered tail cone characteristic of Boeing aircraft, differentiating it from Airbus's straighter-top tail design.
    \end{enumerate}
  \end{tcolorbox}
  \vspace{4pt}
  \begin{tcolorbox}[qwenbox]
    \begin{enumerate}[leftmargin=*,topsep=0pt,itemsep=2pt,parsep=0pt]
      \item To identify the MD-11, look for four under-wing pylon-mounted engines, a T-tail configuration, and a pointed nose profile.
      \item To distinguish from similar variants, focus on the longer fuselage compared to other MD models and the presence of winglets.
      \item To avoid misclassification, check for the T-tail design, differentiating the MD-11 from other aircraft variants.
    \end{enumerate}
  \end{tcolorbox}
\end{minipage}

\newpage
\subsection*{Oxford-IIIT Pets}

\noindent
\begin{minipage}[t]{0.32\textwidth}
  \centering
  \includegraphics[width=\linewidth,height=5cm,keepaspectratio]{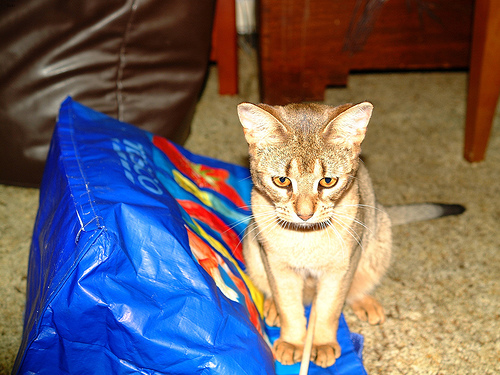}\\[4pt]
  {\small\textit{Ground Truth: \textbf{Abyssinian}}}
  \vspace{8pt}

  \begin{tcolorbox}[gemmabox]
    \begin{enumerate}[leftmargin=*,topsep=0pt,itemsep=2pt,parsep=0pt]
      \item To identify the Abyssinian, look for a ticked tabby coat pattern with distinct agouti hairs.
      \item To distinguish from similar breeds, focus on the wedge-shaped head and almond-shaped eyes.
      \item To avoid misclassification, check for a slender and muscular build, contrasting with the stockier appearance of some domestic shorthairs.
    \end{enumerate}
  \end{tcolorbox}
  \vspace{4pt}
  \begin{tcolorbox}[qwenbox]
    \begin{enumerate}[leftmargin=*,topsep=0pt,itemsep=2pt,parsep=0pt]
      \item To identify the Abyssinian, look for a short, dense coat with a distinctive ticked or spotted pattern.
      \item To distinguish from similar breeds, focus on the rounded head and large, expressive eyes.
      \item To avoid misclassification, check for the absence of long fur and the presence of a distinct ticked pattern across the entire body.
    \end{enumerate}
  \end{tcolorbox}
\end{minipage}
\hfill
\begin{minipage}[t]{0.32\textwidth}
  \centering
  \includegraphics[width=\linewidth,height=5cm,keepaspectratio]{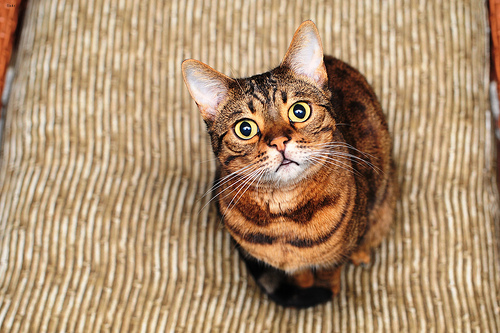}\\[4pt]
  {\small\textit{Ground Truth: \textbf{Bengal}}}
  \vspace{8pt}

  \begin{tcolorbox}[gemmabox]
    \begin{enumerate}[leftmargin=*,topsep=0pt,itemsep=2pt,parsep=0pt]
      \item To identify the Bengal, look for a distinctive spotted or marbled coat pattern with dark brown or black markings.
      \item To distinguish from similar breeds, focus on the overall muscular and athletic build, often exhibiting a ``wildcat'' appearance.
      \item To avoid misclassification, check for a short, dense coat and a relatively small, rounded head compared to other spotted breeds.
    \end{enumerate}
  \end{tcolorbox}
  \vspace{4pt}
  \begin{tcolorbox}[qwenbox]
    \begin{enumerate}[leftmargin=*,topsep=0pt,itemsep=2pt,parsep=0pt]
      \item To identify the Bengal, look for a coat with a distinctive marbled or spotted pattern.
      \item To distinguish from similar breeds, focus on the large, expressive eyes and the athletic, lean body structure.
      \item To avoid misclassification, check for the presence of a tufted tail, which is a key feature of the Bengal breed.
    \end{enumerate}
  \end{tcolorbox}
\end{minipage}
\hfill
\begin{minipage}[t]{0.32\textwidth}
  \centering
  \includegraphics[width=\linewidth,height=5cm,keepaspectratio]{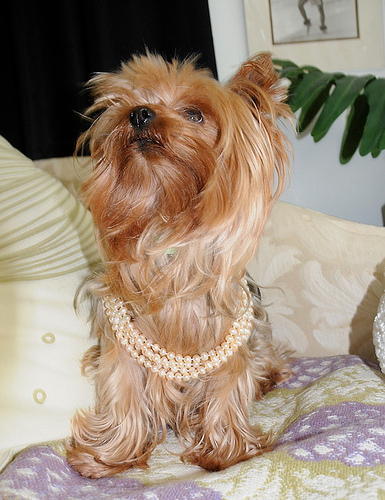}\\[4pt]
  {\small\textit{Ground Truth: \textbf{Yorkshire Terrier}}}
  \vspace{8pt}

  \begin{tcolorbox}[gemmabox]
    \begin{enumerate}[leftmargin=*,topsep=0pt,itemsep=2pt,parsep=0pt]
      \item To identify the Yorkshire Terrier, look for a long, silky, blue-gray or black coat.
      \item To distinguish from similar breeds, focus on the erect ears and a small, compact body.
      \item To avoid misclassification, check for the characteristic lion-clip hairstyle and petite size, differentiating it from larger terrier breeds.
    \end{enumerate}
  \end{tcolorbox}
  \vspace{4pt}
  \begin{tcolorbox}[qwenbox]
    \begin{enumerate}[leftmargin=*,topsep=0pt,itemsep=2pt,parsep=0pt]
      \item To identify the Yorkshire Terrier, look for a small, compact dog with a long, silky coat.
      \item To distinguish from similar breeds, focus on the distinctive pointed ears and expressive eyes.
      \item To avoid misclassification, check for the typical small size and the absence of other breed-specific traits like a large build or different coat texture.
    \end{enumerate}
  \end{tcolorbox}
\end{minipage}

\clearpage
\twocolumn